\documentclass[10pt,twocolumn,letterpaper]{article}

\usepackage[pagenumbers]{iccv} 

\definecolor{iccvblue}{rgb}{0.21,0.49,0.74}
\usepackage[pagebackref,breaklinks,colorlinks,allcolors=iccvblue]{hyperref}

\usepackage[utf8]{inputenc} 
\usepackage[T1]{fontenc}    
\usepackage{hyperref}       
\usepackage{url}            
\usepackage{booktabs}       
\usepackage{amsfonts}       
\usepackage{nicefrac}       
\usepackage{microtype}      
\usepackage{xcolor}         
\usepackage{graphicx}
\usepackage{tabularray}
\usepackage{amsmath}
\usepackage{multirow}
\usepackage{wrapfig}
\usepackage[capitalize]{cleveref}
\usepackage{subcaption}
\usepackage{tabularx}
\usepackage{colortbl}
\usepackage{multicol}
\definecolor{darkgreen}{RGB}{0, 154, 85}
\newcolumntype{Y}{>{\centering\arraybackslash}X}

\definecolor{myred}{RGB}{192, 0, 0}
\definecolor{myyellow}{RGB}{255, 192, 0}
\usepackage{soul}

\newcommand{\localtableofcontents}{
    \begingroup
    \setcounter{tocdepth}{2}
    \hypersetup{linkcolor=black}
    \renewcommand{\baselinestretch}{1.5}
    \setlength{\parskip}{0.06em}
    \tableofcontents
    \endgroup
}

\def\paperID{12928} 
\def\confName{ICCV}
\def\confYear{2025}

\title{V2XPnP: Vehicle-to-Everything Spatio-Temporal Fusion for \\ Multi-Agent Perception and Prediction}

\author{
Zewei Zhou, Hao Xiang, Zhaoliang Zheng, Seth Z. Zhao, Mingyue Lei, Yun Zhang, Tianhui Cai, \\[0.05cm]
Xinyi Liu, Johnson Liu, Maheswari Bajji, Xin Xia, Zhiyu Huang\thanks{Corresponding author. {\tt \{zeweizhou, zhiyuh\}@ucla.edu}}, \ 
Bolei Zhou, Jiaqi Ma\\[0.13cm]
University of California, Los Angeles \\
\small \tt{\href{https://mobility-lab.seas.ucla.edu/v2xpnp/}{\tt https://mobility-lab.seas.ucla.edu/v2xpnp}}
}

\begin{document}
\maketitle

\renewcommand{\thefootnote}{\arabic{footnote}}
\addtocontents{toc}{\protect\setcounter{tocdepth}{-1}}
\begin{abstract}
Vehicle-to-everything (V2X) technologies offer a promising paradigm to mitigate the limitations of constrained observability in single-vehicle systems. Prior work primarily focuses on single-frame cooperative perception, which fuses agents' information across different spatial locations but ignores temporal cues and temporal tasks (e.g., temporal perception and prediction). In this paper, we focus on the spatio-temporal fusion in V2X scenarios and design one-step and multi-step communication strategies (when to transmit) as well as examine their integration with three fusion strategies - early, late, and intermediate (what to transmit), providing comprehensive benchmarks with 11 fusion models (how to fuse). Furthermore, we propose \textbf{V2XPnP}, a novel intermediate fusion framework within one-step communication for end-to-end perception and prediction. Our framework employs a unified Transformer-based architecture to effectively model complex spatio-temporal relationships across multiple agents, frames, and high-definition maps. Moreover, we introduce the \textbf{V2XPnP Sequential Dataset} that supports all V2X collaboration modes and addresses the limitations of existing real-world datasets, which are restricted to single-frame or single-mode cooperation. Extensive experiments demonstrate that our framework outperforms state-of-the-art methods in both perception and prediction tasks. 
\end{abstract}    
\section{Introduction}
\label{sec:intro}

Autonomous driving systems are required to accurately perceive surrounding road users and predict their future trajectories to ensure safe and interactive driving. Despite recent advances in perception and prediction~\cite{gao2025stamp, hu_planning-oriented_2023, song2024collaborative}, single-vehicle systems still struggle with limited perception range and occlusion issues \cite{xu_opv2v_2022, li_v2x-sim_2022}, compromising driving performance and road safety. Consequently, vehicle-to-everything (V2X) technologies have emerged as a promising paradigm to address these challenges, which enable connected and automated vehicles (CAVs) and infrastructures to share complementary information and mitigate occlusions, thereby supporting holistic environment understanding \cite{huang2024v2x, yangjie2024towards, han2024foundation}.

\begin{figure}[t]
    \centering
    \includegraphics[width=0.97\columnwidth]{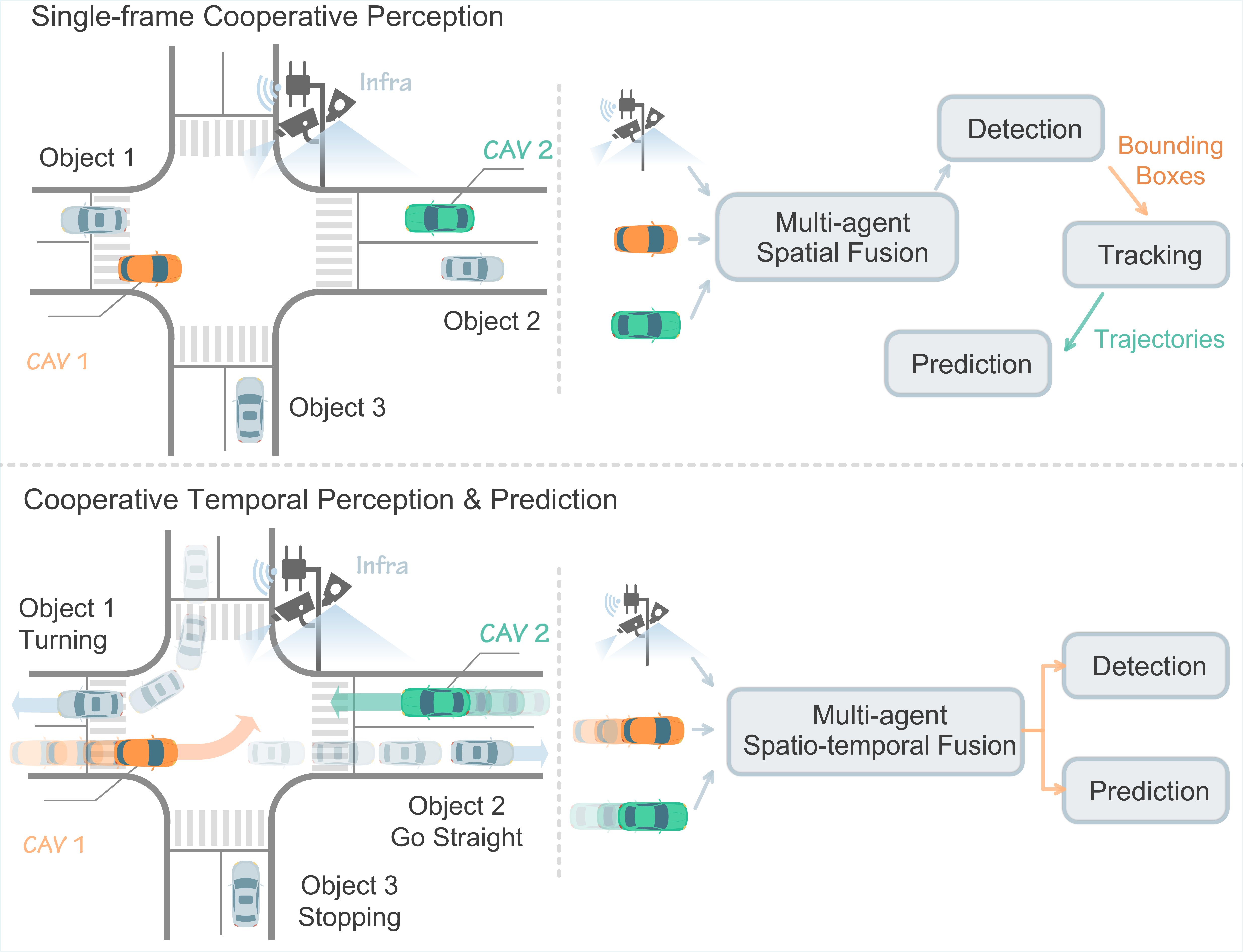}
    \vspace{-0.15cm}
    \caption{Illustration of V2X temporal tasks and our V2X spatio-temporal fusion framework. By incorporating temporal information, our framework enhances V2X communication and supports end-to-end perception and prediction beyond single-frame perception.}
    \label{fig:idea}
    \vspace{-0.45cm}
\end{figure}

Despite their potential, existing works focus on frame-by-frame cooperative detection \cite{lu2024extensible, xiang_hm-vit_2023, zimmer2024tumtraf, li2022multi}, which aggregates information from agents at different spatial locations. However, these works overlook temporal cues across sequential frames, which are important for locating previously visible but currently undetected objects \cite{yang_spatio-temporal_2023} and predicting object future trajectories \cite{luo_fast_2018}. Although some work \cite{ffnet, wei2023asynchronyrobust} incorporate short-term temporal cues (0.5 s) in single-frame perception to mitigate asynchrony, the broader challenge of efficiently aggregating multi-agent and multi-frame information and supporting long-term temporal tasks, such as motion prediction, remains largely unexplored. Therefore, we aim to address critical questions in multi-agent multi-frame cooperation: \textit{(1) What information to transmit? (2) When to transmit it? (3) How to fuse information across multi-agent spatial and temporal dimensions?} 
To address \textbf{\textit{what to transmit}}, we expand the three fusion strategies in single-frame cooperative perception (\ie, early, late, and intermediate) to incorporate the temporal dimension. Regarding \textbf{\textit{when to transmit}}, we introduce one-step and multi-step communication strategies to capture multi-frame temporal information. For \textbf{\textit{how to fuse}}, we conduct a systematic analysis across various spatio-temporal fusion strategies, providing comprehensive benchmarks for cooperative perception and prediction tasks across all V2X collaboration modes. 

Among these strategies, we advocate intermediate fusion within one-step communication because it effectively balances the trade-off between accuracy and increased transmission load. Moreover, its capability to transmit intermediate spatial-temporal features makes it well-suited for end-to-end perception and prediction, allowing for feature sharing across multiple tasks, as shown in \cref{fig:idea}. Based on this strategy, we propose \textit{\textbf{V2XPnP}}, a V2X spatio-temporal fusion framework leveraging a unified Transformer structure for effective spatial-temporal fusion, encompassing temporal attention, self-spatial attention, multi-agent spatial attention, and map attention. Each agent first extracts its inter-frame and self-spatial features, which can support single-vehicle perception and prediction while reducing the communication load, and then the multi-agent spatial attention model fuses the single-agent features across different agents.

Another challenge is the lack of real-world sequential datasets encompassing diverse V2X collaboration modes. In V2X scenarios, vehicles and infrastructures serve as primary agents, with collaboration modes that include vehicle-to-vehicle (V2V), vehicle-to-infrastructure (V2I), and infrastructure-to-infrastructure (I2I), vehicle-centric (VC), and infrastructure-centric (IC) \cite{yangjie2024towards}. Most existing datasets \cite{xiang2024v2x, yu_dair-v2x_2022} are non-sequential, limited to single collaboration modes, and focus only on single-frame cooperative perception, lacking support for temporal tasks. To bridge this gap, we introduce the first large-scale real-world \textbf{\textit{V2XPnP Sequential Dataset}}, featuring four agents and supporting all collaboration modes. This dataset includes temporally consistent perception and trajectory data across 49 2V+2I scenarios, 38 V2V scenarios, and 9 V+2I scenarios, totaling 40k frames, along with point-cloud and vector maps from 24 collected intersections. The main contributions of this paper can be summarized as follows: 
\begin{enumerate}
\item We present \textit{V2XPnP}, a V2X spatio-temporal fusion framework with a novel intermediate fusion model within one-step communication. This framework is based on a unified Transformer architecture integrating diverse attention fusion modules for V2X spatial-temporal information.
\item We introduce the first large-scale, real-world V2X sequential dataset featuring multiple agents and all V2X collaboration modes (\ie, VC, IC, V2V, I2I), encompassing perception data, object trajectories, and map data.
\item We conduct extensive analysis across various spatio-temporal fusion strategies and benchmarks 11 fusion models for cooperative perception and prediction in all V2X collaboration modes, demonstrating the state-of-the-art performance of the proposed model.
\end{enumerate}
\section{Related Work}
\label{sec:RelatedWork}

\begin{figure*}[t]
    \centering
    \includegraphics[width=0.93\textwidth]{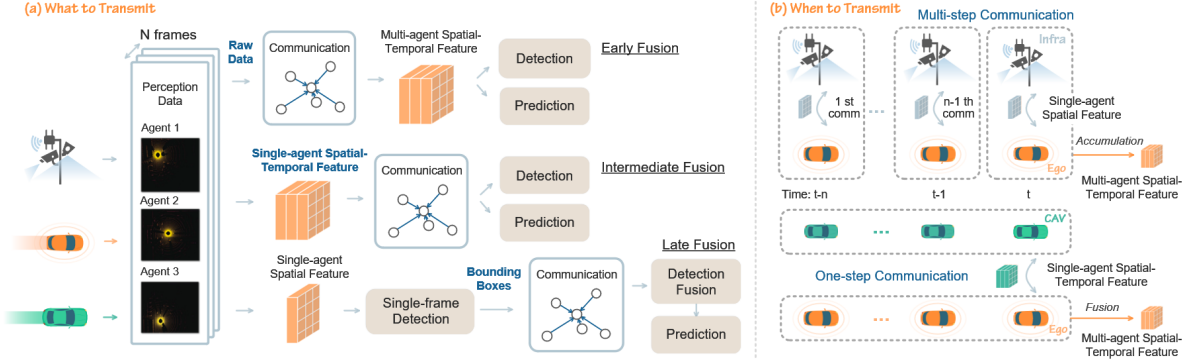}
    \vspace{-0.15cm}
    \caption{Illustration of various V2X fusion strategies for perception and prediction. (a) \textit{What to Transmit}: early, intermediate, and late fusion, transmitting raw sensor data, intermediate BEV features, or bounding boxes. (b) \textit{When to Transmit}: multi-step and one-step communication.}
    \label{fig:fusion strategies}
    \vspace{-0.45cm}
\end{figure*}

\textbf{End-to-end Perception and Prediction.} 
Safe autonomous driving fundamentally depends on accurate perception and prediction \cite{huang_survey_2022, coopre}. In single-agent systems, significant efforts have been dedicated to temporal perception and prediction \cite{li_bevformer_2022}, leading to the development of various end-to-end frameworks. These frameworks enhance computational efficiency by sharing information across tasks and mitigating error propagation inherent in modular architectures. FaF \cite{luo_fast_2018} and PnPNet \cite{liang_pnpnet_2020} focus on LiDAR-based joint perception and prediction, while occupancy flow methods \cite{agro2024uno, agro2023implicit} provide detailed spatial-temporal information. Recently, Bird's-Eye-View (BEV)-based approaches with camera data have gained prominence \cite{gu_vip3d_2023, hu_planning-oriented_2023, jiang2023vad}.

\noindent \textbf{V2X Perception and Prediction.} 
V2X perception has been extensively explored, with intermediate fusion emerging as a widely adopted strategy \cite{xu_opv2v_2022,xu_v2x-vit_2022, gao2025stamp}. FFNet \cite{ffnet} and CoBEVFlow \cite{wei2023asynchronyrobust} utilize historical information (0.5s) from collaborators to mitigate asynchrony, and SCOPE integrates ego-history for detection. However, coordinating multi-agent systems for long-term temporal tasks remains an open challenge. 
In the prediction domain, deep learning models have been extensively studied for modeling inter-agent interactions \cite{shi_mtr_2023, jia_hdgt_2023}. However, the limited short-term visibility of individual vehicles continues to restrict prediction accuracy. Cooperative prediction leveraging V2X has shown potential, though research remains preliminary \cite{zhang2025co, wang2024cmp, yang_spatio-temporal_2023, ruan_learning_2023}. 

To integrate perception and prediction within an end-to-end framework, V2VNet \cite{wang_v2vnet_2020} employs a graph neural network for spatio-temporal fusion. UniV2X \cite{yu2024end} extended the end-to-end system to support downstream tasks, but with a simplified spatio-temporal fusion module. Despite these advancements, a comprehensive framework for V2X-based spatio-temporal fusion is still lacking.

\noindent \textbf{Real-world Driving Datasets.}  
Public datasets have been instrumental in advancing autonomous driving research. Early sequential datasets \cite{noauthor_next_nodate, krajewski_highd_2018} provided only object trajectories on highways \cite{zhou_comprehensive_2022}, but lack perception data. The following datasets, such as nuScenes \cite{caesar_nuscenes_2020} and Waymo \cite{sun_scalability_2020}, introduced real-world urban data but were limited to single-agent perspectives, rendering them unsuitable for V2X research. Thus, simulated datasets like V2XSet \cite{xu_v2x-vit_2022} were developed. Recently, datasets including Dair-V2X \cite{yu_dair-v2x_2022}, V2V4Real \cite{xu_v2v4real_2023}, RCooper \cite{hao2024rcooper}, and V2X-Real \cite{xiang2024v2x} significant contributed to real-world data in the V2I, V2V, I2I and V2X modes. However, real-world sequential V2X datasets covering all collaboration modes remain scarce. V2X-Seq \cite{yu_v2x-seq_2023} is the only sequential dataset incorporated with various behavior and map data for prediction tasks; however, is limited to V2I data and has restricted accessibility with download constraints.
\section{V2XPnP Fusion Framework}

The cooperative temporal perception and prediction task requires the integration of temporal information across historical frames and spatial information from multiple agents. It is defined as follows: given map and historical sequences of raw perception data ${\mathbf{P}_i^t}, i\in\{1,\cdots,N\}$, from all $N$ agents within the communication range of the ego agent, the objective is to detect objects in the current frame and predict their future trajectories considering the map information.

\subsection{V2X Spatio-Temporal Fusion Strategies}

This section addresses what and when to transmit for spatio-temporal fusion, and \cref{section: fusion pipeline} delves into the fusion process.

\noindent \textbf{What to Transmit.} 
In multi-agent spatial fusion, three fusion strategies are widely adopted in single-frame cooperative perception, \ie, early, late, and intermediate fusion \cite{hu2022where2comm, yu_dair-v2x_2022}, which involve the transmission of raw perception data, bounding boxes, and intermediate features, respectively. Adopting this framework, we extend these fusion approaches to the V2X spatio-temporal fusion context, as illustrated in \cref{fig:fusion strategies}.  (1) \textit{Early fusion} transmits the entire raw historical perception data to retain complete feature information but imposes the highest transmission load. (2) \textit{Late fusion} shares only the final detected results at each historical frame, resulting in the lowest transmission load but losing most of the feature information. (3) \textit{Intermediate fusion} transmits intermediate spatio-temporal BEV features, striking a balance between information quality and transmission load.

\noindent \textbf{When to Transmit.} 
Determining when to transmit in spatio-temporal fusion is more challenging due to the inclusion of temporal data as compared to the existing spatial fusion. A straightforward approach is \textit{Multi-step Communication}, where each transmission only includes the current frame’s data. 
However, the multiple steps cause the accumulation of delays and data loss, and obtaining complete historical data requires that neighboring agents stay within the ego agent's limited communication range throughout historical frames. 
In practice, the ego agent should aggregate as much data as possible from other agents within a single transmission, rather than relying on multiple exchanges to obtain complete spatio-temporal information. To address this, we propose a \textit{One-step Communication} strategy, where individual agents share all their historical data within a single communication. 

\noindent \textbf{Intermediate Fusion with One-step Communication.} 
Synthesizing the considerations for \textit{what to transmit} and \textit{when to transmit}, we adopt an intermediate fusion within the one-step communication strategy, which fuses temporal data from multiple frames before transmission, allowing each agent to share aggregated information without excessive communication overhead. In multi-step intermediate fusion, each agent transmits BEV feature maps at every timestep $\mathbf{F}_{i}^{t} \in \mathbb{R}^{H \times W \times C}$, which denotes agent $i$'s BEV feature at time $t$ with height $H$, width $W$, and channels $C$. The cumulative data shared across $T$ frames is a stacked sequence $\mathbf{F}_{i}^{seq}\in \mathbb{R}^{T \times H \times W \times C}$, resulting in a significantly higher transmission load than single-frame, along with potential delays and data loss. Conversely, in our proposed strategy, each agent first fuses its historical BEV features internally, reducing the sequence from $\mathbf{F}_{i}^{seq}$ to a single condensed BEV feature map $\mathbf{F}^{'}_{i}\in \mathbb{R}^{H \times W \times C}$. This reduction allows agents to transmit a compact data packet, comparable in size to single-frame cooperative perception, thereby conserving bandwidth while preserving essential spatio-temporal information.

\begin{figure*}[t]
    \centering
    \includegraphics[width=0.95\textwidth]{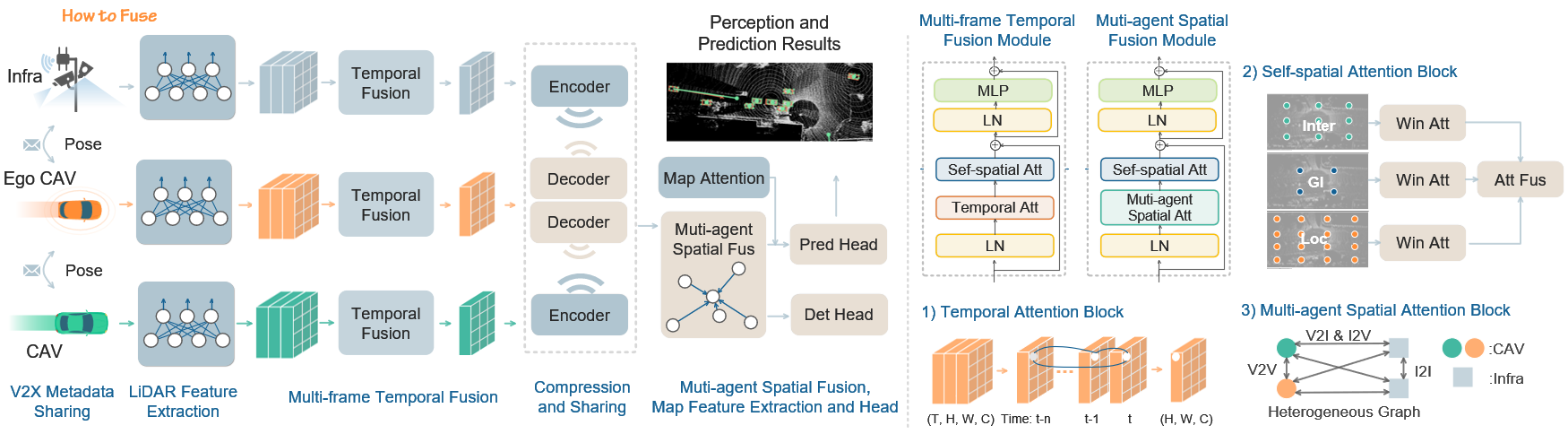}
    \vspace{-0.15cm}
    \caption{The V2XPnP framework and multi-agent spatio-temporal fusion model. The framework comprises various components for feature extraction, fusion, and decoding. Within our fusion model, we introduce multiple attention mechanisms to enhance the fusion process.}
    \label{fig:framework and model}
    \vspace{-0.45cm}
\end{figure*}

\subsection{V2XPnP Framework}
\label{section: fusion pipeline}
The spatio-temporal features of intermediate fusion render it a natural fit for end-to-end perception and prediction. Accordingly, we propose a unified end-to-end perception and prediction framework to perform multiple tasks across spatio-temporal dimensions. The overall V2XPnP framework is illustrated in \cref{fig:framework and model}, which includes six components and is unfolded in this section. The detail of the spatio-temporal fusion model is provided in \cref{sec:ST transformer}. Notably, each module in V2XPnP is modular, allowing for easy replacement.

\noindent \textbf{V2X Metadata Sharing.}
Each agent in the V2X system is an observer and collaborator in the shared environment. Each agent will first determine its collaborators with its communication range, and share the metadata, such as relative poses and extrinsic, to construct a spatial V2X graph. Each node and edge in the graph represents an agent and a communication channel. All the point clouds are transformed into the ego agent's coordinate frame before feature extraction.

\noindent \textbf{LiDAR Feature Extraction.}  
We utilize the PointPillar network \cite{lang_pointpillars_2019} to extract the LiDAR feature for each agent $i$ at time $t$, which has low inference latency. The extracted features are structured into a 2D pseudo-image representation to produce salient feature maps $\mathbf{F}_{i}^{t}$, and the sequential feature is stacked as $\mathbf{F}^{seq}_{i} \in \mathbb{R}^{T \times H \times W \times C}$

\noindent \textbf{Multi-frame Temporal Fusion.} 
We propose a Transformer-based temporal fusion module to iteratively perform inter-frame and intra-agent BEV feature fusion through self-attention mechanisms. The spatio-temporal feature of each agent is extracted through the temporal dimension while minimizing communication overhead. The feature map for each agent after temporal fusion is $\mathbf{F}^{'}_{i} \in \mathbb{R}^{H \times W \times C}$.

\noindent \textbf{Compression and Sharing.} 
To reduce the transmission load, intermediate features are compressed using a $1 \times 1$ kernel convolution network in its channel dimension. The ego agent uses another convolution network to decompress the features, restoring them to their original dimensionality.

\noindent \textbf{Multi-agent Spatial Fusion.} 
The decompressed features are passed into a Transformer-based multi-agent spatial fusion network to learn inter-agent and intra-agent spatio-temporal interaction and update the multi-agent feature map $\mathbf{F}^{'}$.

\noindent \textbf{Map Feature Extraction.} 
The HD map is directly accessed for each agent without V2X fusion. We project the vectorized HD map to BEV space by incorporating the map polylines into each BEV feature grid. We first employ a multi-layer perceptron ($\operatorname{MLP}$) to encode the surrounding map polylines for each grid $\mathbf{M}\in\mathbb{R}^{H\times W\times N_{m} \times n \times D}$, resulting in the map feature $\mathbf{F}_{m} \in \mathbb{R}^{H \times W \times N_{m} \times C}$. Here, $N_{m}$ and $n$ represent the number of map polylines and the waypoints per polyline, while $D$ represents waypoint attributes (\ie, position and lane type). The map encoding is expressed as: $\mathbf{F}_{m}=\phi\left(\operatorname{MLP}\left(\mathbf{M}\right)\right)$, where $\phi$ denotes max-pooling on the waypoint axis. Then, a map-BEV attention is introduced to inject the map feature into the BEV feature. We concatenate the BEV and map feature $\mathbf{F}_{bm}=[\mathbf{F}^{'}, \mathbf{F}_{m}] \in \mathbb{R}^{H \times W \times (1+N_{m}) \times C}$, and add the position embedding $\mathbf{P}_{m}$ based on sinusoidal positional encoding. The final content feature map $\mathbf{F} \in \mathbb{R}^{H \times W \times C}$ is updated by multi-head self-attention ($\operatorname{MHSA}$) as follows:
\begin{equation}
\mathbf{F}\!=\!\operatorname{MHSA}\bigl(\operatorname{Q}\colon[\mathbf{F}_{bm},\!\mathbf{P}_{m}], \operatorname{K}\colon[\mathbf{F}_{bm},\!\mathbf{P}_{m}], \operatorname{V}\colon\mathbf{F}_{bm}\bigr).
\end{equation}

\noindent \textbf{Detection and Prediction Heads.}
Finally, a detection head and a prediction head are connected to the final feature $\mathbf{F}$ to output states for each predefined anchor box: (1) The detection head contains two 2D convolution layers for bounding box regression and classification. The regression branch outputs the bounding box position $(x,y,z)$, size $(w,l,h)$, and yaw $\theta$. The classification branch outputs the confidence score of each anchor box, determining whether it corresponds to an object or background. (2) The prediction head outputs offset values for each anchor box at each future timestamp using two 2D convolution layers, and the final trajectory is generated by accumulating these offsets.

\subsection{Spatio-Temporal Fusion Transformer}
\label{sec:ST transformer}

In this section, we introduce spatio-temporal fusion with a unified Transformer architecture. The proposed model comprises three blocks: temporal attention, self-spatial attention, and multi-agent spatial attention, as shown in \cref{fig:framework and model}, and two core fusion modules. (1) \textit{Multi-frame temporal fusion}: Each agent first extracts their spatio-temporal features through iterative temporal and self-spatial attentions. (2) \textit{Multi-agent spatial fusion}: rich BEV features from multiple agents are acquired via V2X and then fused through iterative multi-agent spatial and self-spatial attentions.

\noindent\textbf{Temporal Attention.} 
This block is designed to capture the inter-frame relationship and aggregate historical BEV features $\mathbf{F}^{seq}_{i}$ across the temporal dimension. The history timestamps are encoded with a learnable embedding, which is added to each BEV feature frame to form $\mathbf{F}^{seq'}_{i}$. To preserve temporal cues, this block only fuses temporal features from the same spatial positions across frames, and spatial features are further extracted by the self-spatial Transformer. Temporal fusion is expressed as:
\begin{align}
\mathbf{F}_{i}^{tem} &= \operatorname{MHSA}\bigl(\operatorname{Q}\colon\operatorname{MLP}(\mathbf{F}^{seq'}_{i}), \operatorname{K}\colon\operatorname{MLP}(\mathbf{F}^{seq'}_{i}), \nonumber \\
&\quad\quad\quad\quad\: \: \operatorname{V}\colon\operatorname{MLP}(\mathbf{F}^{seq'}_{i})\bigr).
\end{align}

\noindent \textbf{Self-spatial Attention.} 
To capture the intra-agent spatial BEV interaction, this block employs multi-scale window attention to capture spatial features at different resolutions and ranges. A large window focuses on global features for long-term behavior, and a small window preserves local finer information. Note that this block only fuses spatial features for each agent in a frame without inter-frame and inter-agent fusion. Specifically, we utilize local, intermediate, and global windows $P_{k}\in\{P_{loc}, P_{inter}, P_{gl}\}$ to partition the feature map through the $H$ and $W$ dimension, generating the self-spatial token $\mathbf{F}^{sp}\in\mathbb{R}^{\frac{H}{P_k} \times \frac{W}{P_k} \times {P_k}^2 \times C}$. With an additional relative position encoding, $\operatorname{MHSA}$ is operated among ${P_k}^2$ tokens. The final output is obtained by performing split attention to fuse features from different windows.

\noindent \textbf{Multi-agent Spatial Attention.} 
This block facilitates inter-agent fusion by aggregating BEV feature maps from multiple agents. Considering the different deployment positions and capabilities of vehicle and infrastructure sensors, the multi-agent spatial Transformer is heterogeneous with individual learnable weights for different interaction pairs (\ie, V-I, V-V, I-V, and I-I). The attention token of $i$ th agent $\mathbf{F}_{i, m}^{sp}$ is modulated by its type $m$ embedding, and is weighted by the relation matrix $\mathbf{W}_{\text{att}}^{(e_{i,j})}$ between edge $e_{i,j}$ during aggregation with agent $j$ of type $n$:

\begin{align}
&\mathbf{Q}_i^m\! =\! \operatorname{MLP}(\mathbf{F}_{i,m}),  \mathbf{K}_j^n\! =\! \operatorname{MLP}(\mathbf{F}_{j,n}),  \mathbf{V}_j^n\! =\! \operatorname{MLP}(\mathbf{F}_{j,n}), \nonumber \\
&\mathbf{F}_{i,m}^{sp} = \sum_j \operatorname{Softmax}( \mathbf{Q}_i^m \cdot \mathbf{W}_{\text{att}}^{(e_{i,j})} \cdot \mathbf{K}_j^n ) \cdot \mathbf{V}_j^n.
\end{align}

\subsection{Learning Objective}
The learning objective is comprised of temporal perception and prediction tasks. First, we define the perception loss as the combination of regression $\mathcal{L}_{\text{reg}}$ and classification loss $\mathcal{L}_{\text{cla}}$ of the predefined anchor box. Specifically, the smooth $\ell_{1}$-loss is leveraged for the regression part, and the focal loss \cite{lin_focal_2017} is utilized for classification. Second, we define the prediction loss $\mathcal{L}_{\text {pred }}$ as $\ell_2$-loss between the prediction points sequence with ground truth trajectory. The final loss function is the weighted sum of $\mathcal{L}_{\text {reg }}, \mathcal{L}_{\text {cla }}$, and $\mathcal{L}_{\text {pred }}$.

\begin{figure*}[t]
    \centering
    \includegraphics[width=0.95\textwidth]{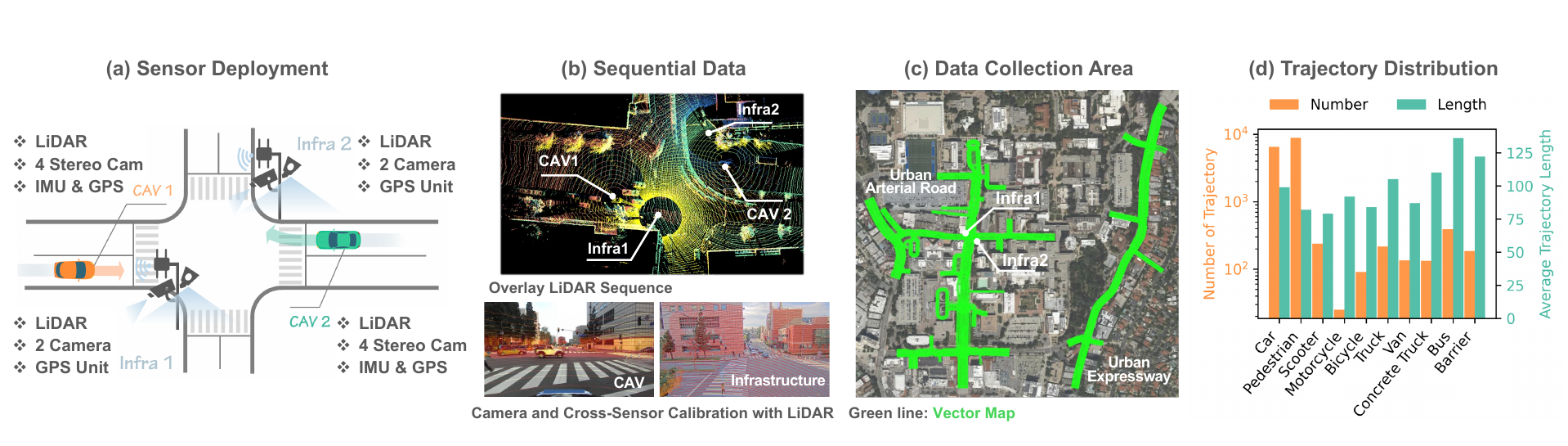}
    \vspace{-0.4cm}
    \caption{Illustration of the V2XPnP Sequential Dataset. (a) V2X data acquisition systems; (b) Sequential LiDAR and camera data; (c) Data collection area and vector map; (d) Total number and average tracking length of 3D tracked objects per category.}
    \label{fig:dataset}
    \vspace{-0.5cm}
\end{figure*}

\section{V2XPnP Sequential Dataset}
We introduce the V2XPnP-Sequential dataset, the first large-scale, real-world V2X sequential dataset featuring multiple agents and all collaboration modes. This dataset comprises 96 scenarios (49 2V+2I scenarios, 38 V2V scenarios, and 9 V+2I scenarios), each spanning 95 to 283 frames captured at 10 Hz. The dataset comprises two data sequences from CAV perception (point clouds and camera images) and two data sequences from infrastructure perception, as shown in \cref{fig:dataset}(b). We also provide corresponding vector maps and point-cloud maps for all collection areas, as shown in \cref{fig:dataset}(c). Ten object categories are included, and the average trajectory length and frequency of each category are shown in \cref{fig:dataset}(d). Further details on data visualization, annotation, trajectory and map generation are provided in the supplementary.

\subsection{Data Acquisition}
V2X temporal tasks require diverse time-consistent perception data and object behavior data. We choose urban arterial roads, expressways, and intersections to collect V2X data. The sensor configurations for two CAVs and two infrastructures are shown in \cref{fig:dataset}(a). By permuting and combining the behavior patterns of CAVs (such as overtaking, platooning, turning, etc.), we designed a total of about 60 interaction pairs between CAVs to collect data. From more than 66h driving logs, we annotate 28 representative scenarios. We also process the non-sequential data from V2X-Real \cite{xiang2024v2x} with a V2X sequential data processing pipeline, detailed in \cref{sec:seq_process}. Since perception tasks do not need to consider labeling consistency, identical objects may be assigned different IDs within a sequence, leading to data fragmentation and limiting support for temporal tasks. The final dataset is yielded by processing our collected data and V2X-Real data.

\subsection{Sequential Data Processing}
\label{sec:seq_process}

Time-consistent data is crucial for temporal tasks, and thus we develop a V2X sequential data processing pipeline to track objects across time and different agents' views. We construct a multi-agent spatio-temporal graph, where each node $n_{it}^k$ represents an annotated bounding box $i$ at time $t$ from agent $k$’s LiDAR data, and edges connect nodes that correspond to the same object. Identifying objects with the same ID then reduces to finding connected components in this graph, with each isolated component representing a unique object. Moreover, we build this graph by leveraging the temporal continuity in single-agent annotations and incorporating multi-agent data. For temporal consistency within a single agent, we add edges between $n_{it}^k$ and $n_{it+1}^k$, although annotation errors may sometimes assign different IDs to the same object. To address this, we integrate annotations from multiple agents, transforming annotations into a global coordinate frame and connecting nodes if their Intersection-over-Union (IoU) exceeds a threshold. Once the graph is complete, we identify connected components and assign each a unique tracking ID. To mitigate annotation biases, we refine object attributes based on their consensus.

\section{Experiments}
\subsection{Experimental Setup}

\textbf{Evaluation Metrics}.
Following the detection evaluation protocol in \cite{xiang2024v2x}, we measure detection performance using Average Precision (AP) at an IoU threshold of 0.5. For prediction, we provide the results of commonly used metrics \cite{trajectron++, densetnt, saadatnejad2022sattack}, including Average Displacement Error (ADE), Final Displacement Error (FDE), and Miss Rate (MR) within a 2-meter threshold. However, prediction accuracy is influenced by false positives and missed detections from the perception module. For example, a poorly performing perception module that detects only a simple object in a straight-line trajectory can misleadingly inflate the prediction accuracy. To address this, we employ the \textit{End-to-end Perception and Prediction Accuracy} (EPA) metric \cite{gu_vip3d_2023} to jointly evaluate perception and prediction performance.
\begin{equation}
\vspace{-0.3cm}
\text{EPA} = \frac{|\hat{S}_{\text{match, hit}}| - \alpha N_{\text{FP}}}{N_{\text{GT}}},
\vspace{-0.1cm}
\end{equation}
where $|\hat{S}_{\text{match, hit}}|$ is the number of true positive objects with prediction $FDE <\tau_{EPA}$, $N_{FP}, N_{GT}$ represent the number of false positive objects and ground truth objects, respectively, and $\alpha$ is a penalty coefficient. A higher EPA value indicates superior object detection and prediction capabilities, and we set $\tau_{EPA}=2$m, $\alpha=0.5$ following \cite{gu_vip3d_2023}.

\begin{table*}[t]
\centering
\caption{Benchmark results of cooperative perception and prediction models on V2XPnP Sequential (V2XPnP-Seq) Dataset}
\label{tab:benchmarks}
\vspace{-0.1cm}
\scriptsize
\renewcommand{\arraystretch}{1.11}
\setlength{\tabcolsep}{9.6pt}
\begin{tabularx}{0.95\textwidth}{@{}l|l|cc|c|ccc|c@{}}
\toprule
Dataset & Method & E2E & Map & \textbf{AP@0.5} (\%) $\uparrow$ & ADE (m) $\downarrow$ & FDE (m) $\downarrow$ & MR (\%) $\downarrow$ & \textbf{EPA} (\%) $\uparrow$ \\ \midrule
\multirow{8}{*}{\begin{tabular}[c]{@{}l@{}}V2XPnP-Seq-VC\\ {\scriptsize \textit{(with V+2I at most)}}\end{tabular}}  
    & No Fusion & & \checkmark &43.9  &1.87  &3.24  &33.8  &24.3  \\
    & No Fusion-FaF* \cite{luo_fast_2018} &\checkmark & &53.4 &1.55 &2.81 &34.3 &31.6 \\ 
    & Late Fusion & & \checkmark &58.1  &1.59  &2.82  &32.4  &33.0  \\
    & Early Fusion & \checkmark & \checkmark &60.3  &1.37  &2.49  &33.8  & 36.7 \\ \cmidrule(l){2-9} 
    & CoBEVFlow* \cite{wei2023asynchronyrobust} & \checkmark & \checkmark &63.3 &1.36  &2.49 &33.0  &41.9   \\
    & FFNet* \cite{ffnet} & \checkmark & \checkmark &64.6  &1.36  &2.47  &34.7  &42.3  \\
    & V2X-ViT* \cite{xu_v2x-vit_2022} & \checkmark & \checkmark &69.6  &1.39  &2.56  &35.2  &44.7  \\
    & \cellcolor{gray!20}V2XPnP (Ours) & \cellcolor{gray!20}\checkmark & \cellcolor{gray!20}\checkmark & \cellcolor{gray!20} \textbf{71.6} \tiny{\textcolor{darkgreen}{+2.0}}& \cellcolor{gray!20} {1.35} &\cellcolor{gray!20} {2.36} &\cellcolor{gray!20} {31.7} &\cellcolor{gray!20} \textbf{48.2} \tiny{\textcolor{darkgreen}{+3.5}}\\ \midrule

\multirow{8}{*}{\begin{tabular}[c]{@{}l@{}}V2XPnP-Seq-IC\\ {\scriptsize \textit{(with 2V+I at most)}}\end{tabular}}  
    & No Fusion & & \checkmark &46.4 &1.69  &3.06  & 36.2 &28.8  \\
    & No Fusion-FaF* \cite{luo_fast_2018} & \checkmark & &56.7  &1.34  &2.65  & 41.4 &31.7    \\
    & Late Fusion & & \checkmark &55.9  &1.39  &2.44  &{30.1} &32.9  \\
    & Early Fusion & \checkmark & \checkmark &60.5  &1.39  &2.63  &32.8  &39.5  \\ \cmidrule(l){2-9} 
    & CoBEVFlow* \cite{wei2023asynchronyrobust} & \checkmark & \checkmark &57.6 &1.38  &2.58  &31.0  &32.5   \\
    & FFNet* \cite{ffnet} & \checkmark & \checkmark &61.0  &1.18  &2.18  &35.1  &37.5  \\
    & V2X-ViT* \cite{xu_v2x-vit_2022} & \checkmark & \checkmark &69.3  &1.27  &2.39  &35.4  &43.3  \\
    & \cellcolor{gray!20}V2XPnP (Ours) & \cellcolor{gray!20}\checkmark & \cellcolor{gray!20}\checkmark & \cellcolor{gray!20}\textbf{71.0} \tiny{\textcolor{darkgreen}{+1.7}}& \cellcolor{gray!20}{1.18} & \cellcolor{gray!20}{2.16} & \cellcolor{gray!20}{34.0} & \cellcolor{gray!20}\textbf{46.0} \tiny{\textcolor{darkgreen}{+2.7}}\\ \midrule

\multirow{8}{*}{V2XPnP-Seq-V2V}  
    & No Fusion & & \checkmark &40.8  &1.99  &3.38  &34.0  &19.8  \\
    & No Fusion-FaF* \cite{luo_fast_2018} & \checkmark & &51.9  &1.67  &3.12  &39.3  &27.5  \\
    & Late Fusion & & \checkmark &55.3  &1.75  &{3.07}  &{34.0}  &30.5  \\
    & Early Fusion & \checkmark & \checkmark &53.0  &{1.64}  &3.11  &40.2  &26.9 \\ \cmidrule(l){2-9} 
    & CoBEVFlow* \cite{wei2023asynchronyrobust} & \checkmark & \checkmark &58.7 &1.72  &3.15  &40.3  &33.6   \\
    & FFNet* \cite{ffnet} & \checkmark & \checkmark &56.5  &1.68  &3.12  &39.8  &31.2  \\
    & V2X-ViT* \cite{xu_v2x-vit_2022} & \checkmark & \checkmark &64.6  &1.68  &3.13  &39.8  &36.7  \\
    & \cellcolor{gray!20}V2XPnP (Ours) & \cellcolor{gray!20}\checkmark & \cellcolor{gray!20}\checkmark & \cellcolor{gray!20}\textbf{70.5} \tiny{\textcolor{darkgreen}{+5.9}}& \cellcolor{gray!20}{1.78} & \cellcolor{gray!20}{3.28} & \cellcolor{gray!20}{39.9} & \cellcolor{gray!20}\textbf{40.6} \tiny{\textcolor{darkgreen}{+3.9}} \\ \midrule

\multirow{8}{*}{V2XPnP-Seq-I2I}  
    & No Fusion & & \checkmark &51.0  &1.69  &3.06  &36.2  &31.7  \\
    & No Fusion-FaF* \cite{luo_fast_2018} & \checkmark & &56.6 &1.34  &2.65  &41.4  &31.7  \\
    & Late Fusion & & \checkmark &61.3  &1.41  &2.50  &{30.0}  &41.6  \\
    & Early Fusion & \checkmark & \checkmark &64.6  &1.57  &2.98  &39.9  &37.7  \\ \cmidrule(l){2-9} 
    & CoBEVFlow* \cite{wei2023asynchronyrobust} & \checkmark & \checkmark &58.4 &1.31  &2.61  &41.5  &33.0   \\
    & FFNet* \cite{ffnet} & \checkmark & \checkmark &66.1  &1.41  &2.59  &36.3  &40.9  \\
    & V2X-ViT* \cite{xu_v2x-vit_2022} & \checkmark & \checkmark &65.4  &{1.22}  &2.33  &35.9  &41.3  \\
    & \cellcolor{gray!20}V2XPnP (Ours) & \cellcolor{gray!20}\checkmark & \cellcolor{gray!20}\checkmark & \cellcolor{gray!20}\textbf{69.2} \tiny{\textcolor{darkgreen}{+3.1}} & \cellcolor{gray!20}{1.26} & \cellcolor{gray!20}{2.31} & \cellcolor{gray!20}{36.5} & \cellcolor{gray!20}\textbf{42.8} \tiny{\textcolor{darkgreen}{+1.2}}\\ \bottomrule
\end{tabularx}%
\vspace{-0.3cm}
\end{table*}

\noindent \textbf{Collaboration Modes}.
The V2XPnP Sequential Dataset supports various V2X collaboration modes by organizing data with specific interaction patterns. \textit{Vehicle-Centric (VC)}: The ego CAV is the focal agent, communicating with other CAVs and infrastructure (Infra). \textit{Infrastructure-Centric (IC)}: Infrastructure is the central entity, communicating with other Infras and CAVs. \textit{Vehicle-to-Vehicle (V2V)}: The ego CAV communicates exclusively with other CAVs without involving Infra. \textit{Infrastructure-to-Infrastructure (I2I)}: Infra shares data only with each other. Each VC and IC scenario includes 2-4 agents, which is close to real-world V2X settings and can evaluate model generalization across diverse V2X scenarios, whereas each V2V and I2I scenario has two agents.

\noindent \textbf{Implementation Details}. 
During the testing stage, we select a fixed agent as the ego agent in each cooperative scenario, while the ego agent is shuffled and randomly selected during training. Following the real-time setting \cite{cooperfuse}, we set the communication range to 50 meters and evaluate surrounding agents within a range of $x\in[- 70, 70]$m and $y\in[-40,40]$m. Messages beyond 50 meters are discarded. Besides, the history length is 2 s (2 Hz), and the prediction horizon is 3 s (2 Hz). The train/validation/test data splits are 76/6/14 scenarios. Additional training and model details are provided in the supplementary materials.


\subsection{Benchmark Methods}
\noindent \textbf{End-to-end methods.} Most of the single-frame perception models cannot support the prediction task, thus, we first implement a baseline end-to-end model with the same LiDAR backbone and decoding heads as V2XPnP but utilizing the temporal fusion module FaF \cite{luo_fast_2018} - alternating 2D and 3D convolutions - as the \textit{No fusion-FaF$^*$} baseline, which can extend the non-temporal model to support temporal tasks. Then, the early fusion configurations and several state-of-the-art intermediate fusion models are integrated as: \textit{Early Fusion}, \textit{FFNet}$^*$ \cite{ffnet}, \textit{CoBEVFlow}$^*$ \cite{wei2023asynchronyrobust}, \textit{V2X-ViT}$^*$ \cite{xu_v2x-vit_2022}). More benchmark results of \textit{DiscoNet}$^*$ \cite{disconet}, \textit{F-Cooper}$^*$ \cite{chen2019f}, and \textit{V2VNet}$^*$ \cite{wang_v2vnet_2020} are provided in the supplementary. These models marked with $^*$ are reimplemented in our framework with the same LiDAR backbone and decoding heads.

\noindent \textbf{Decoupled methods.} 
Transmitting final detection results renders late fusion incompatible with end-to-end models. Thus, we benchmark \textit{Late Fusion} with a decoupled pipeline, where objects in each historical frame are detected using a single-frame perception module, and results are fused via non-maximum suppression. Assuming an ideal tracker to generate object trajectories from perception results and interpolate missing points, we implement an attention-based predictor for trajectory-level prediction tasks, following the prediction mainstream \cite{shi_mtr_2023}. To further assess end-to-end performance, we also evaluate a decoupled \textit{No-Fusion} model.

\subsection{Results}

\cref{tab:benchmarks} presents the benchmark results across four V2X collaboration modes. Since prediction performance inherently depends on detection quality, the difficulty posed by different detected objects can significantly impact prediction accuracy. Thus, \ul{\textit{the EPA metric serves as the most appropriate indicator for assessing the overall performance}}. More details and baseline results are provided in the supplementary.

\noindent \textbf{What to Transmit.} In both end-to-end and decoupled PnP frameworks, the cooperation perception and prediction performance are consistently better than non-fusion models in all collaboration modes, especially in the primary EPA metric, which demonstrates the benefits of cooperation in temporal perception and prediction. Moreover, intermediate fusion models (\eg, V2X-ViT, FFNet, and V2XPnP) generally outperform other fusion strategies, while early fusion consistently surpasses late fusion. Our proposed V2XPnP model achieves the best performance, outperforming other competitive cooperative methods.

\begin{figure*}[t]
    \centering
    \includegraphics[width=0.99\textwidth]{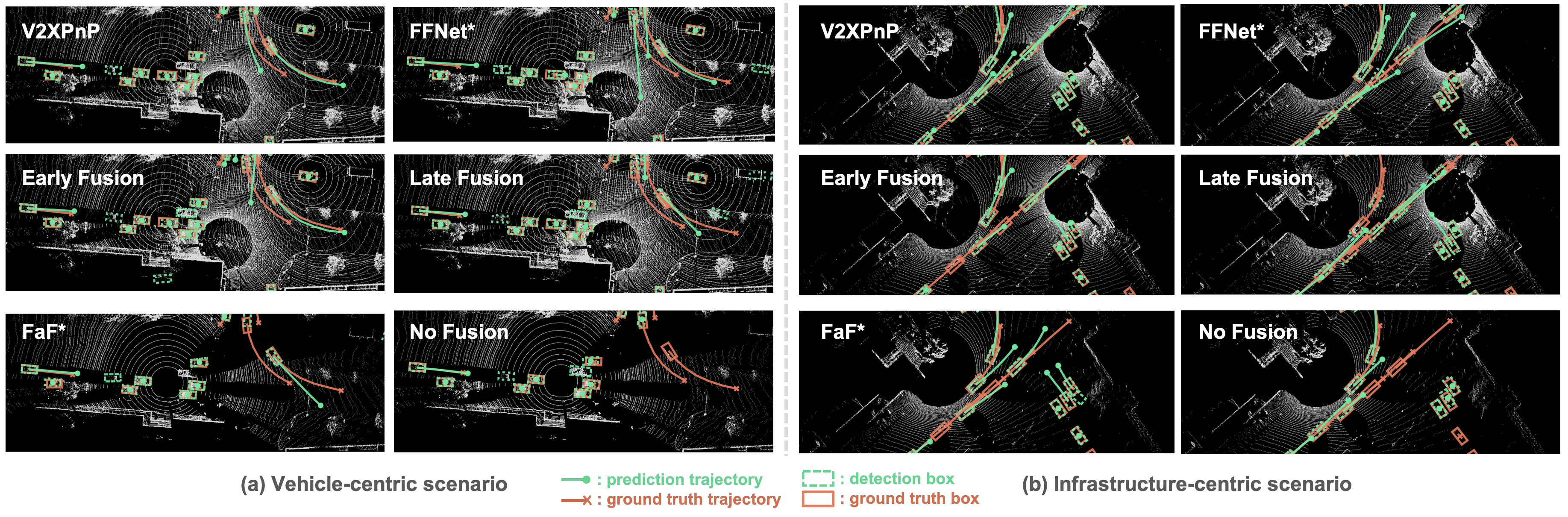}
    \vspace{-0.25cm}
    \caption{Qualitative results of different fusion models on the testing set. V2XPnP shows better perception and prediction results.}
    \label{fig:qua_results}
    \vspace{-0.45cm}
\end{figure*}

\begin{table}[t]
\caption{Comparison of one-step and multi-step communication}
\vspace{-0.2cm}
\label{tab:communication strategies}
\setlength{\tabcolsep}{7pt} 
\footnotesize 
\begin{tabularx}{\columnwidth}{l|l|lll|l}
\toprule
Strategy                         & \textbf{AP@0.5} $\uparrow$     & ADE $\downarrow$     & FDE $\downarrow$     & MR $\downarrow$       & \textbf{EPA} $\uparrow$ \\ \midrule
Multi-step                          &68.2                       &1.56                      &2.84                      &31.8                       & 43.0                         \\
\rowcolor{gray!20} One-step                        &71.6                   &1.35                           &2.36                   &31.7                   & 48.2 
                       \\ \bottomrule
\end{tabularx}
\vspace{-0.3cm}
\end{table}

\begin{table}[t]
\caption{Ablation results of V2XPnP model}
\vspace{-0.2cm}
\label{tab:Ablation}
\footnotesize
\setlength{\tabcolsep}{3.8pt} 
\begin{tabularx}{\columnwidth}{ccc|c|ccc|c}
\toprule
{Temp} &
{Spatial} &
{Map} &
\textbf{AP@0.5} $\uparrow$ &
{ADE} $\downarrow$ &
{FDE} $\downarrow$ &
{MR} $\downarrow$ &
\textbf{EPA} $\uparrow$ \\ \midrule
                          &            &            & 43.9          & -    & -    & -    & -    \\
\checkmark                &            &            & 57.2          & 1.52 & 2.76 & 35.5 & 33.8 \\
\checkmark                & \checkmark &            & 71.3          & 1.48 & 2.70 & 36.2 & 44.4 \\
\rowcolor{gray!20}\checkmark & \checkmark & \checkmark & \textbf{71.6} & \textbf{1.35} & \textbf{2.36} & \textbf{31.7} & \textbf{48.2} \\ 
\bottomrule
\end{tabularx}
\vspace{-0.4cm} 
\end{table}

\noindent \textbf{When to Transmit.} 
\cref{tab:communication strategies} shows the performance of our model under two communication strategies. We find that the perception and prediction performance of one-step communication is improved compared to multi-step communication by 5.0\% AP and 12\% in EPA. This improvement arises because each agent first directly fuses their lossless raw temporal data before sharing, avoiding error accumulation from lossy intermediate information transformed in multi-step communication across temporal dimensions. Moreover, the one-step strategy compresses spatio-temporal feature transmission (under a 32$\times$ compression rate) from 5$\times$0.269 Mb to 0.269 Mb compared to multi-step communication, while mitigating information loss when agents move out of communication range in historical frames. At a typical C-V2X data transmission rate \cite{arena2019overview}, the transmission delay of one-step communication is approximately $10\sim20$ ms.

\noindent \textbf{How to Fuse.} 
\cref{tab:Ablation} provides the ablation study of V2XPnP, showing the effectiveness of different components in V2XPnP. The temporal fusion module provides the history information for current-frame detection, while multi-agent spatial fusion alleviates occlusions and improves performance by incorporating other views. The map fusion module enhances trajectory prediction by guiding future trajectories to align with road structures. Our complete V2XPnP model with all these fusion modules performs the best.

\noindent \textbf{End-to-end vs. Decoupled Frameworks.} 
The end-to-end model consistently outperforms the decoupled framework. In no-fusion situations, the end-to-end model FaF$^*$ outperforms the decoupled model in detection by leveraging temporal cues. Furthermore, FaF$^*$ achieves performance comparable to late fusion with V2X spatial aggregation due to integrating temporal features. The intermediate fusion of spatio-temporal features in V2XPnP aligns well with the end-to-end architecture, showcasing its superior performance.

\noindent \textbf{Infrastructure vs. Vehicle Centric.} 
As shown in \cref{tab:benchmarks}, models under VC and IC modes outperform those in V2V and I2I modes, because VC and IC can aggregate information from up to four agents rather than only two, resulting in enhanced environmental understanding. Notably, the evaluation protocol is consistent across all models within each mode. Additionally, stationary infrastructure-based agents in IC and I2I modes offer higher prediction accuracy by providing elevated sensing perspectives and less noisy data.

\begin{figure}[t]
    \centering
    \includegraphics[width=\columnwidth]{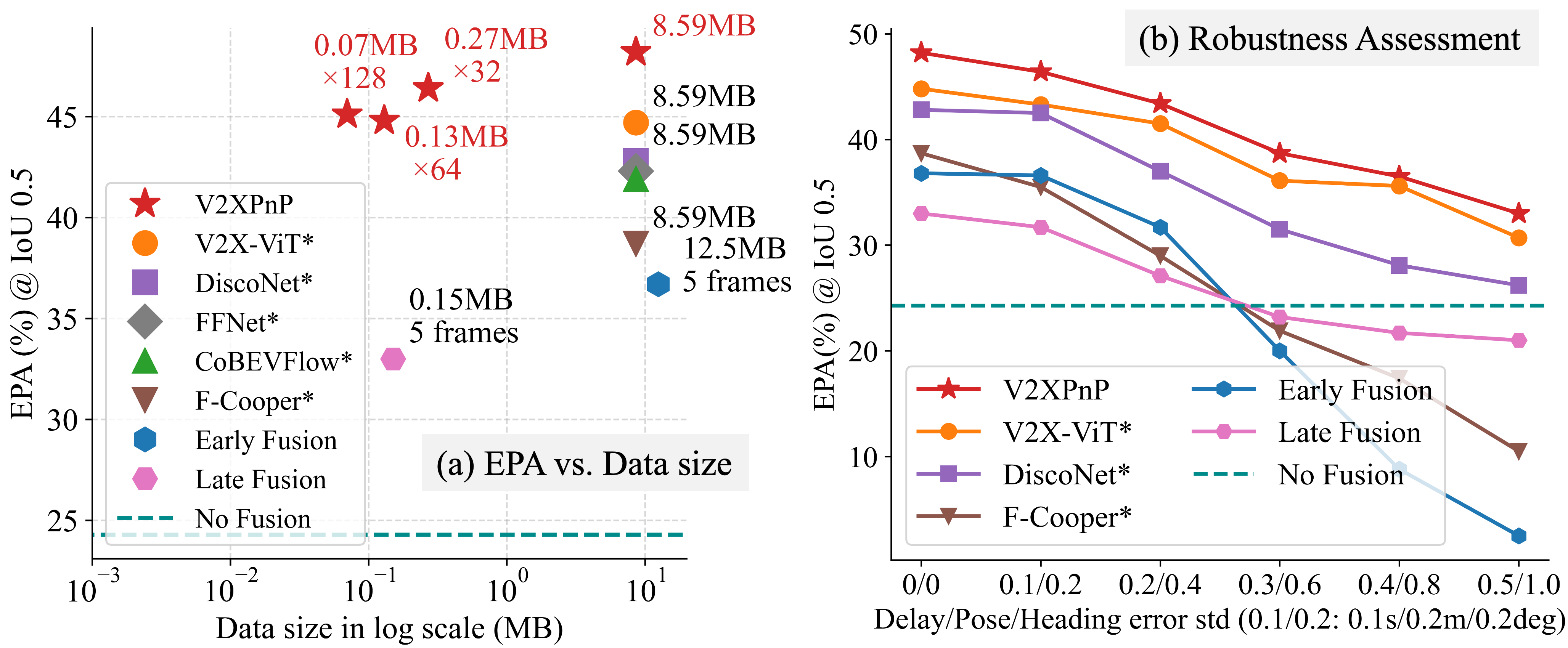}
    \vspace{-0.6cm}
    \caption{Transmission data size and communication noise test results. V2XPnP shows better performance with varying data compression rates and robustness under communication noise.}
    \label{fig:delay and compression}
    \vspace{-0.5cm}
\end{figure}

\noindent \textbf{Transmission Data Size and Robustness Test.}
Results in \cref{fig:delay and compression} indicate that V2XPnP achieves a good balance between communication efficiency and accuracy, maintaining superior performance compared to full-size V2X-ViT* even at a $\times 128$ compression rate. Following the setting in \cite{xu_v2x-vit_2022}, we provide the results with 100-500 ms time delay and positional/head Gaussian noise from (0.2m, 0.2°) to (1m, 1°). Both V2XPnP and V2X-ViT* maintain robust performance due to their spatial attention fusion, and V2XPnP performs better due to its designed temporal attention.

\noindent \textbf{Qualitative Results.}
\cref{fig:qua_results} visualizes the outcomes of cooperative perception and prediction across different fusion models. The No Fusion model is constrained by its limited field of view. The FaF model, leveraging temporal information within an end-to-end pipeline, performs better under occlusion. Late and early fusion models significantly benefit from multi-agent data integration, though late fusion remains impacted by error propagation, such as detection heading errors misleading trajectory direction. Notably, the end-to-end intermediate fusion model, particularly V2XPnP, performs better in both detection and prediction tasks.

\section{Conclusions}
We propose V2XPnP, a novel V2X spatio-temporal fusion framework for cooperative temporal perception and prediction. The core of this framework is a unified Transformer-based model for spatio-temporal fusion and map fusion.  Furthermore, we examine various fusion strategies concerning what, when to transmit, and how to fuse, offering comprehensive benchmarks. Additionally, we introduce the V2X Sequential Dataset, which supports all V2X collaboration modes. Extensive experiments demonstrate the superior performance of the proposed framework, establishing its efficacy in advancing cooperative temporal tasks. 

\section*{Acknowledgements}
This work was supported by the Federal Highway Administration Center of Excellence on New Mobility and Automated Vehicles, and by the National Science Foundation under Award No. 2346267, POSE: Phase II - DriveX: An Open Source Ecosystem for Automated Driving and Intelligent Transportation Research.

{\small
\bibliographystyle{ieeenat_fullname}
\bibliography{main}

\begin{thebibliography}{57}
\providecommand{\natexlab}[1]{#1}
\providecommand{\url}[1]{\texttt{#1}}
\expandafter\ifx\csname urlstyle\endcsname\relax
  \providecommand{\doi}[1]{doi: #1}\else
  \providecommand{\doi}{doi: \begingroup \urlstyle{rm}\Url}\fi

\bibitem[noa()]{noauthor_next_nodate}
Next {Generation} {Simulation} ({NGSIM}) {Vehicle} {Trajectories} and {Supporting} {Data} {\textbar} {USDOT} {Open} {Data}.

\bibitem[Agro et~al.(2023)Agro, Sykora, Casas, and Urtasun]{agro2023implicit}
Ben Agro, Quinlan Sykora, Sergio Casas, and Raquel Urtasun.
\newblock Implicit occupancy flow fields for perception and prediction in self-driving.
\newblock In \emph{Proceedings of the IEEE/CVF Conference on Computer Vision and Pattern Recognition}, pages 1379--1388, 2023.

\bibitem[Agro et~al.(2024)Agro, Sykora, Casas, Gilles, and Urtasun]{agro2024uno}
Ben Agro, Quinlan Sykora, Sergio Casas, Thomas Gilles, and Raquel Urtasun.
\newblock Uno: Unsupervised occupancy fields for perception and forecasting.
\newblock In \emph{Proceedings of the IEEE/CVF Conference on Computer Vision and Pattern Recognition}, pages 14487--14496, 2024.

\bibitem[Arena and Pau(2019)]{arena2019overview}
Fabio Arena and Giovanni Pau.
\newblock An overview of vehicular communications.
\newblock \emph{Future internet}, 11\penalty0 (2):\penalty0 27, 2019.

\bibitem[Caesar et~al.(2020)Caesar, Bankiti, Lang, Vora, Liong, Xu, Krishnan, Pan, Baldan, and Beijbom]{caesar_nuscenes_2020}
Holger Caesar, Varun Bankiti, Alex~H. Lang, Sourabh Vora, Venice~Erin Liong, Qiang Xu, Anush Krishnan, Yu Pan, Giancarlo Baldan, and Oscar Beijbom.
\newblock nuscenes: {A} multimodal dataset for autonomous driving.
\newblock In \emph{Proceedings of the {IEEE}/{CVF} conference on computer vision and pattern recognition}, pages 11621--11631, 2020.

\bibitem[Chen et~al.(2019)Chen, Ma, Tang, Guo, Yang, and Fu]{chen2019f}
Qi Chen, Xu Ma, Sihai Tang, Jingda Guo, Qing Yang, and Song Fu.
\newblock F-cooper: Feature based cooperative perception for autonomous vehicle edge computing system using 3d point clouds.
\newblock In \emph{Proceedings of the 4th ACM/IEEE Symposium on Edge Computing}, pages 88--100, 2019.

\bibitem[Crescenzi et~al.(2001)Crescenzi, Mecca, Merialdo, et~al.]{crescenzi2001roadrunner}
Valter Crescenzi, Giansalvatore Mecca, Paolo Merialdo, et~al.
\newblock Roadrunner: Towards automatic data extraction from large web sites.
\newblock In \emph{VLDB}, pages 109--118, 2001.

\bibitem[Deo and Trivedi(2018)]{deo_convolutional_2018}
Nachiket Deo and Mohan~M. Trivedi.
\newblock Convolutional {Social} {Pooling} for {Vehicle} {Trajectory} {Prediction}.
\newblock In \emph{2018 {IEEE}/{CVF} {Conference} on {Computer} {Vision} and {Pattern} {Recognition} {Workshops} ({CVPRW})}, pages 1549--15498, Salt Lake City, UT, USA, 2018. IEEE.

\bibitem[Ettinger et~al.(2021)Ettinger, Cheng, Caine, Liu, Zhao, Pradhan, Chai, Sapp, Qi, Zhou, et~al.]{ettinger2021large}
Scott Ettinger, Shuyang Cheng, Benjamin Caine, Chenxi Liu, Hang Zhao, Sabeek Pradhan, Yuning Chai, Ben Sapp, Charles~R Qi, Yin Zhou, et~al.
\newblock Large scale interactive motion forecasting for autonomous driving: The waymo open motion dataset.
\newblock In \emph{Proceedings of the IEEE/CVF International Conference on Computer Vision}, pages 9710--9719, 2021.

\bibitem[Gao et~al.(2025)Gao, Xu, Li, Wang, Fan, and Tu]{gao2025stamp}
Xiangbo Gao, Runsheng Xu, Jiachen Li, Ziran Wang, Zhiwen Fan, and Zhengzhong Tu.
\newblock Stamp: Scalable task and model-agnostic collaborative perception.
\newblock \emph{arXiv preprint arXiv:2501.18616}, 2025.

\bibitem[Gu et~al.(2021)Gu, Sun, and Zhao]{densetnt}
Junru Gu, Chen Sun, and Hang Zhao.
\newblock Densetnt: End-to-end trajectory prediction from dense goal sets.
\newblock In \emph{Proceedings of the IEEE/CVF International Conference on Computer Vision}, pages 15303--15312, 2021.

\bibitem[Gu et~al.(2023)Gu, Hu, Zhang, Chen, Wang, Wang, and Zhao]{gu_vip3d_2023}
Junru Gu, Chenxu Hu, Tianyuan Zhang, Xuanyao Chen, Yilun Wang, Yue Wang, and Hang Zhao.
\newblock {ViP3D}: {End}-to-{End} {Visual} {Trajectory} {Prediction} via {3D} {Agent} {Queries}.
\newblock 2023.

\bibitem[Han et~al.(2024)Han, Meng, Xia, Liao, He, Zheng, Wang, Xiang, Zhou, Gao, et~al.]{han2024foundation}
Xu Han, Zonglin Meng, Xin Xia, Xishun Liao, Yueshuai He, Zhaoliang Zheng, Yutong Wang, Hao Xiang, Zewei Zhou, Letian Gao, et~al.
\newblock Foundation intelligence for smart infrastructure services in transportation 5.0.
\newblock \emph{IEEE Transactions on Intelligent Vehicles}, 2024.

\bibitem[Hao et~al.(2024)Hao, Fan, Dai, Zhang, Li, Wang, Yu, Yang, Jirui, and Nie]{hao2024rcooper}
Ruiyang Hao, Siqi Fan, Yingru Dai, Zhenlin Zhang, Chenxi Li, Yuntian Wang, Haibao Yu, Wenxian Yang, Yuan Jirui, and Zaiqing Nie.
\newblock Rcooper: A real-world large-scale dataset for roadside cooperative perception.
\newblock In \emph{Proceedings of the IEEE/CVF Conference on Computer Vision and Pattern Recognition (CVPR)}, 2024.

\bibitem[Hu et~al.(2022)Hu, Fang, Lei, Zhong, and Chen]{hu2022where2comm}
Yue Hu, Shaoheng Fang, Zixing Lei, Yiqi Zhong, and Siheng Chen.
\newblock Where2comm: Communication-efficient collaborative perception via spatial confidence maps.
\newblock \emph{Advances in neural information processing systems}, 35:\penalty0 4874--4886, 2022.

\bibitem[Hu et~al.(2023)Hu, Yang, Chen, Li, Sima, Zhu, Chai, Du, Lin, Wang, Lu, Jia, Liu, Dai, Qiao, and Li]{hu_planning-oriented_2023}
Yihan Hu, Jiazhi Yang, Li Chen, Keyu Li, Chonghao Sima, Xizhou Zhu, Siqi Chai, Senyao Du, Tianwei Lin, Wenhai Wang, Lewei Lu, Xiaosong Jia, Qiang Liu, Jifeng Dai, Yu Qiao, and Hongyang Li.
\newblock Planning-oriented {Autonomous} {Driving}, 2023.
\newblock arXiv:2212.10156 [cs].

\bibitem[Huang et~al.(2024)Huang, Wang, Xia, Chen, Yang, Wang, and Wen]{huang2024v2x}
Xun Huang, Jinlong Wang, Qiming Xia, Siheng Chen, Bisheng Yang, Cheng Wang, and Chenglu Wen.
\newblock V2x-r: Cooperative lidar-4d radar fusion for 3d object detection with denoising diffusion.
\newblock \emph{arXiv preprint arXiv:2411.08402}, 2024.

\bibitem[Huang et~al.(2022)Huang, Du, Yang, Zhou, Zhang, and Chen]{huang_survey_2022}
Yanjun Huang, Jiatong Du, Ziru Yang, Zewei Zhou, Lin Zhang, and Hong Chen.
\newblock A {Survey} on {Trajectory}-{Prediction} {Methods} for {Autonomous} {Driving}.
\newblock \emph{IEEE Transactions on Intelligent Vehicles}, pages 1--1, 2022.
\newblock Conference Name: IEEE Transactions on Intelligent Vehicles.

\bibitem[Ji et~al.(2024)Ji, Zhou, Yang, Yanjun, Yuanjian, Wanting, Lu, and Yu]{yangjie2024towards}
Yangjie Ji, Zewei Zhou, Ziru Yang, Huang Yanjun, Zhang Yuanjian, Zhang Wanting, Xiong Lu, and Zhuoping Yu.
\newblock Towards autonomous vehicles: a survey on cooperative vehicle-infrastructure system.
\newblock \emph{Iscience}, 2024.

\bibitem[Jia et~al.(2023)Jia, Wu, Chen, Liu, Li, and Yan]{jia_hdgt_2023}
Xiaosong Jia, Penghao Wu, Li Chen, Yu Liu, Hongyang Li, and Junchi Yan.
\newblock {HDGT}: {Heterogeneous} {Driving} {Graph} {Transformer} for {Multi}-{Agent} {Trajectory} {Prediction} {Via} {Scene} {Encoding}.
\newblock \emph{IEEE Transactions on Pattern Analysis and Machine Intelligence}, pages 1--15, 2023.
\newblock Conference Name: IEEE Transactions on Pattern Analysis and Machine Intelligence.

\bibitem[Jiang et~al.(2023)Jiang, Chen, Xu, Liao, Chen, Zhou, Zhang, Liu, Huang, and Wang]{jiang2023vad}
Bo Jiang, Shaoyu Chen, Qing Xu, Bencheng Liao, Jiajie Chen, Helong Zhou, Qian Zhang, Wenyu Liu, Chang Huang, and Xinggang Wang.
\newblock Vad: Vectorized scene representation for efficient autonomous driving.
\newblock In \emph{Proceedings of the IEEE/CVF International Conference on Computer Vision}, pages 8340--8350, 2023.

\bibitem[Kingma and Ba(2014)]{kingma2014adam}
Diederik~P Kingma and Jimmy Ba.
\newblock Adam: A method for stochastic optimization.
\newblock \emph{arXiv preprint arXiv:1412.6980}, 2014.

\bibitem[Krajewski et~al.(2018)Krajewski, Bock, Kloeker, and Eckstein]{krajewski_highd_2018}
Robert Krajewski, Julian Bock, Laurent Kloeker, and Lutz Eckstein.
\newblock The {highD} {Dataset}: {A} {Drone} {Dataset} of {Naturalistic} {Vehicle} {Trajectories} on {German} {Highways} for {Validation} of {Highly} {Automated} {Driving} {Systems}.
\newblock In \emph{2018 21st {International} {Conference} on {Intelligent} {Transportation} {Systems} ({ITSC})}, pages 2118--2125, 2018.
\newblock ISSN: 2153-0017.

\bibitem[Lang et~al.(2019)Lang, Vora, Caesar, Zhou, Yang, and Beijbom]{lang_pointpillars_2019}
Alex~H. Lang, Sourabh Vora, Holger Caesar, Lubing Zhou, Jiong Yang, and Oscar Beijbom.
\newblock Pointpillars: {Fast} encoders for object detection from point clouds.
\newblock In \emph{Proceedings of the {IEEE}/{CVF} conference on computer vision and pattern recognition}, pages 12697--12705, 2019.

\bibitem[Li et~al.(2020)Li, Wang, Li, Li, Wu, and Hao]{li2020sustech}
E Li, Shuaijun Wang, Chengyang Li, Dachuan Li, Xiangbin Wu, and Qi Hao.
\newblock Sustech points: A portable 3d point cloud interactive annotation platform system.
\newblock In \emph{2020 IEEE Intelligent Vehicles Symposium (IV)}, pages 1108--1115. IEEE, 2020.

\bibitem[Li et~al.()Li, Zhang, Ma, Wang, and Feng]{li2022multi}
Yiming Li, Juexiao Zhang, Dekun Ma, Yue Wang, and Chen Feng.
\newblock Multi-robot scene completion: Towards task-agnostic collaborative perception.
\newblock In \emph{6th Annual Conference on Robot Learning}.

\bibitem[Li et~al.(2021)Li, Ren, Wu, Chen, Feng, and Zhang]{disconet}
Yiming Li, Shunli Ren, Pengxiang Wu, Siheng Chen, Chen Feng, and Wenjun Zhang.
\newblock Learning distilled collaboration graph for multi-agent perception.
\newblock In \emph{Thirty-fifth Conference on Neural Information Processing Systems (NeurIPS 2021)}, 2021.

\bibitem[Li et~al.(2022{\natexlab{a}})Li, Ma, An, Wang, Zhong, Chen, and Feng]{li_v2x-sim_2022}
Yiming Li, Dekun Ma, Ziyan An, Zixun Wang, Yiqi Zhong, Siheng Chen, and Chen Feng.
\newblock {V2X}-{Sim}: {Multi}-{Agent} {Collaborative} {Perception} {Dataset} and {Benchmark} for {Autonomous} {Driving}.
\newblock \emph{IEEE Robotics and Automation Letters}, 7\penalty0 (4):\penalty0 10914--10921, 2022{\natexlab{a}}.
\newblock Conference Name: IEEE Robotics and Automation Letters.

\bibitem[Li et~al.(2022{\natexlab{b}})Li, Wang, Li, Xie, Sima, Lu, Qiao, and Dai]{li_bevformer_2022}
Zhiqi Li, Wenhai Wang, Hongyang Li, Enze Xie, Chonghao Sima, Tong Lu, Yu Qiao, and Jifeng Dai.
\newblock {BEVFormer}: {Learning} {Bird}’s-{Eye}-{View} {Representation} from {Multi}-camera {Images} via {Spatiotemporal} {Transformers}.
\newblock In \emph{Computer {Vision} – {ECCV} 2022}, pages 1--18, Cham, 2022{\natexlab{b}}. Springer Nature Switzerland.

\bibitem[Liang et~al.(2020)Liang, Yang, Zeng, Chen, Hu, Casas, and Urtasun]{liang_pnpnet_2020}
Ming Liang, Bin Yang, Wenyuan Zeng, Yun Chen, Rui Hu, Sergio Casas, and Raquel Urtasun.
\newblock {PnPNet}: {End}-to-{End} {Perception} and {Prediction} {With} {Tracking} in the {Loop}.
\newblock pages 11553--11562, 2020.

\bibitem[Lin et~al.(2017)Lin, Goyal, Girshick, He, and Dollár]{lin_focal_2017}
Tsung-Yi Lin, Priya Goyal, Ross Girshick, Kaiming He, and Piotr Dollár.
\newblock Focal loss for dense object detection.
\newblock In \emph{Proceedings of the {IEEE} international conference on computer vision}, pages 2980--2988, 2017.

\bibitem[Lu et~al.(2024)Lu, Hu, Zhong, Wang, Wang, and Chen]{lu2024extensible}
Yifan Lu, Yue Hu, Yiqi Zhong, Dequan Wang, Yanfeng Wang, and Siheng Chen.
\newblock An extensible framework for open heterogeneous collaborative perception.
\newblock \emph{arXiv preprint arXiv:2401.13964}, 2024.

\bibitem[Luo et~al.(2018)Luo, Yang, and Urtasun]{luo_fast_2018}
Wenjie Luo, Bin Yang, and Raquel Urtasun.
\newblock Fast and {Furious}: {Real} {Time} {End}-to-{End} {3D} {Detection}, {Tracking} and {Motion} {Forecasting} {With} a {Single} {Convolutional} {Net}.
\newblock pages 3569--3577, 2018.

\bibitem[Ruan et~al.(2023)Ruan, Yu, Yang, Fan, Tang, and Nie]{ruan_learning_2023}
Hongzhi Ruan, Haibao Yu, Wenxian Yang, Siqi Fan, Yingjuan Tang, and Zaiqing Nie.
\newblock Learning {Cooperative} {Trajectory} {Representations} for {Motion} {Forecasting}, 2023.
\newblock arXiv:2311.00371 [cs].

\bibitem[Saadatnejad et~al.(2022)Saadatnejad, Bahari, Khorsandi, Saneian, Moosavi-Dezfooli, and Alahi]{saadatnejad2022sattack}
Saeed Saadatnejad, Mohammadhossein Bahari, Pedram Khorsandi, Mohammad Saneian, Seyed-Mohsen Moosavi-Dezfooli, and Alexandre Alahi.
\newblock Are socially-aware trajectory prediction models really socially-aware?
\newblock \emph{Transportation Research Part C: Emerging Technologies}, 141:\penalty0 103705, 2022.

\bibitem[Salzmann et~al.(2020)Salzmann, Ivanovic, Chakravarty, and Pavone]{trajectron++}
Tim Salzmann, Boris Ivanovic, Punarjay Chakravarty, and Marco Pavone.
\newblock Trajectron++: Dynamically-feasible trajectory forecasting with heterogeneous data.
\newblock In \emph{Computer Vision -- ECCV 2020}, pages 683--700, Cham, 2020. Springer International Publishing.

\bibitem[Shi et~al.(2023)Shi, Jiang, Dai, and Schiele]{shi_mtr_2023}
Shaoshuai Shi, Li Jiang, Dengxin Dai, and Bernt Schiele.
\newblock {MTR}++: {Multi}-{Agent} {Motion} {Prediction} with {Symmetric} {Scene} {Modeling} and {Guided} {Intention} {Querying}, 2023.
\newblock arXiv:2306.17770 [cs].

\bibitem[Song et~al.(2024)Song, Liang, Cao, Yan, Zimmer, Gross, Festag, and Knoll]{song2024collaborative}
Rui Song, Chenwei Liang, Hu Cao, Zhiran Yan, Walter Zimmer, Markus Gross, Andreas Festag, and Alois Knoll.
\newblock Collaborative semantic occupancy prediction with hybrid feature fusion in connected automated vehicles.
\newblock In \emph{Proceedings of the IEEE/CVF Conference on Computer Vision and Pattern Recognition}, pages 17996--18006, 2024.

\bibitem[Sun et~al.(2020)Sun, Kretzschmar, Dotiwalla, Chouard, Patnaik, Tsui, Guo, Zhou, Chai, Caine, Vasudevan, Han, Ngiam, Zhao, Timofeev, Ettinger, Krivokon, Gao, Joshi, Zhang, Shlens, Chen, and Anguelov]{sun_scalability_2020}
Pei Sun, Henrik Kretzschmar, Xerxes Dotiwalla, Aurelien Chouard, Vijaysai Patnaik, Paul Tsui, James Guo, Yin Zhou, Yuning Chai, Benjamin Caine, Vijay Vasudevan, Wei Han, Jiquan Ngiam, Hang Zhao, Aleksei Timofeev, Scott Ettinger, Maxim Krivokon, Amy Gao, Aditya Joshi, Yu Zhang, Jonathon Shlens, Zhifeng Chen, and Dragomir Anguelov.
\newblock Scalability in {Perception} for {Autonomous} {Driving}: {Waymo} {Open} {Dataset}.
\newblock pages 2446--2454, 2020.

\bibitem[Wang et~al.(2020)Wang, Manivasagam, Liang, Yang, Zeng, Tu, and Urtasun]{wang_v2vnet_2020}
Tsun-Hsuan Wang, Sivabalan Manivasagam, Ming Liang, Bin Yang, Wenyuan Zeng, James Tu, and Raquel Urtasun.
\newblock {V2VNet}: {Vehicle}-to-{Vehicle} {Communication} for {Joint} {Perception} and {Prediction}.
\newblock In \emph{{arXiv}:2008.07519 [cs]}, 2020.
\newblock 00010 ECC arXiv: 2008.07519.

\bibitem[Wang et~al.(2024)Wang, Wang, Wu, Ma, Li, Qiu, and Li]{wang2024cmp}
Zehao Wang, Yuping Wang, Zhuoyuan Wu, Hengbo Ma, Zhaowei Li, Hang Qiu, and Jiachen Li.
\newblock Cmp: Cooperative motion prediction with multi-agent communication.
\newblock \emph{arXiv preprint arXiv:2403.17916}, 2024.

\bibitem[Wei et~al.(2023)Wei, Wei, Hu, Lu, Zhong, Chen, and Zhang]{wei2023asynchronyrobust}
Sizhe Wei, Yuxi Wei, Yue Hu, Yifan Lu, Yiqi Zhong, Siheng Chen, and Ya Zhang.
\newblock Asynchrony-robust collaborative perception via bird's eye view flow.
\newblock In \emph{Advances in Neural Information Processing Systems}, 2023.

\bibitem[Xiang et~al.(2023)Xiang, Xu, and Ma]{xiang_hm-vit_2023}
Hao Xiang, Runsheng Xu, and Jiaqi Ma.
\newblock {HM}-{ViT}: {Hetero}-modal {Vehicle}-to-{Vehicle} {Cooperative} perception with vision transformer, 2023.
\newblock arXiv:2304.10628 [cs].

\bibitem[Xiang et~al.(2024)Xiang, Zheng, Xia, Xu, Gao, Zhou, Han, Ji, Li, Meng, et~al.]{xiang2024v2x}
Hao Xiang, Zhaoliang Zheng, Xin Xia, Runsheng Xu, Letian Gao, Zewei Zhou, Xu Han, Xinkai Ji, Mingxi Li, Zonglin Meng, et~al.
\newblock V2x-real: a largs-scale dataset for vehicle-to-everything cooperative perception.
\newblock \emph{arXiv preprint arXiv:2403.16034}, 2024.

\bibitem[Xu et~al.(2022{\natexlab{a}})Xu, Xiang, Tu, Xia, Yang, and Ma]{xu_v2x-vit_2022}
Runsheng Xu, Hao Xiang, Zhengzhong Tu, Xin Xia, Ming-Hsuan Yang, and Jiaqi Ma.
\newblock {V2X}-{ViT}: {Vehicle}-to-{Everything} {Cooperative} {Perception} with {Vision} {Transformer}.
\newblock In \emph{{ECCV} 2022}, pages 107--124, Cham, 2022{\natexlab{a}}. Springer Nature Switzerland.

\bibitem[Xu et~al.(2022{\natexlab{b}})Xu, Xiang, Xia, Han, Li, and Ma]{xu_opv2v_2022}
Runsheng Xu, Hao Xiang, Xin Xia, Xu Han, Jinlong Li, and Jiaqi Ma.
\newblock {OPV2V}: {An} {Open} {Benchmark} {Dataset} and {Fusion} {Pipeline} for {Perception} with {Vehicle}-to-{Vehicle} {Communication}.
\newblock In \emph{2022 {International} {Conference} on {Robotics} and {Automation} ({ICRA})}, pages 2583--2589, 2022{\natexlab{b}}.

\bibitem[Xu et~al.(2023)Xu, Xia, Li, Li, Zhang, Tu, Meng, Xiang, Dong, Song, Yu, Zhou, and Ma]{xu_v2v4real_2023}
Runsheng Xu, Xin Xia, Jinlong Li, Hanzhao Li, Shuo Zhang, Zhengzhong Tu, Zonglin Meng, Hao Xiang, Xiaoyu Dong, Rui Song, Hongkai Yu, Bolei Zhou, and Jiaqi Ma.
\newblock {V2V4Real}: {A} {Real}-{World} {Large}-{Scale} {Dataset} for {Vehicle}-to-{Vehicle} {Cooperative} {Perception}.
\newblock pages 13712--13722, 2023.

\bibitem[Yang et~al.(2023)Yang, Yang, Zhang, Li, Liu, Liu, Wang, Sun, and Song]{yang_spatio-temporal_2023}
Kun Yang, Dingkang Yang, Jingyu Zhang, Mingcheng Li, Yang Liu, Jing Liu, Hanqi Wang, Peng Sun, and Liang Song.
\newblock Spatio-{Temporal} {Domain} {Awareness} for {Multi}-{Agent} {Collaborative} {Perception}.
\newblock pages 23383--2339. arXiv, 2023.

\bibitem[Yu et~al.(2022)Yu, Luo, Shu, Huo, Yang, Shi, Guo, Li, Hu, Yuan, and Nie]{yu_dair-v2x_2022}
Haibao Yu, Yizhen Luo, Mao Shu, Yiyi Huo, Zebang Yang, Yifeng Shi, Zhenglong Guo, Hanyu Li, Xing Hu, Jirui Yuan, and Zaiqing Nie.
\newblock {DAIR}-{V2X}: {A} {Large}-{Scale} {Dataset} for {Vehicle}-{Infrastructure} {Cooperative} {3D} {Object} {Detection}.
\newblock pages 21361--21370, 2022.

\bibitem[Yu et~al.(2023)Yu, Yang, Ruan, Yang, Tang, Gao, Hao, Shi, Pan, Sun, Song, Yuan, Luo, and Nie]{yu_v2x-seq_2023}
Haibao Yu, Wenxian Yang, Hongzhi Ruan, Zhenwei Yang, Yingjuan Tang, Xu Gao, Xin Hao, Yifeng Shi, Yifeng Pan, Ning Sun, Juan Song, Jirui Yuan, Ping Luo, and Zaiqing Nie.
\newblock {V2X}-{Seq}: {A} {Large}-{Scale} {Sequential} {Dataset} for {Vehicle}-{Infrastructure} {Cooperative} {Perception} and {Forecasting}.
\newblock arXiv, 2023.
\newblock arXiv:2305.05938 [cs].

\bibitem[Yu et~al.(2024{\natexlab{a}})Yu, Tang, Xie, Mao, Luo, and Nie]{ffnet}
Haibao Yu, Yingjuan Tang, Enze Xie, Jilei Mao, Ping Luo, and Zaiqing Nie.
\newblock Flow-based feature fusion for vehicle-infrastructure cooperative 3d object detection.
\newblock \emph{Advances in Neural Information Processing Systems}, 36, 2024{\natexlab{a}}.

\bibitem[Yu et~al.(2024{\natexlab{b}})Yu, Yang, Zhong, Yang, Fan, Luo, and Nie]{yu2024end}
Haibao Yu, Wenxian Yang, aru Zhong, Zhenwei Yang, Siqi Fan, Ping Luo, and Zaiqing Nie.
\newblock End-to-end autonomous driving through v2x cooperation.
\newblock \emph{arXiv preprint arXiv:2404.00717}, 2024{\natexlab{b}}.

\bibitem[Zhang et~al.(2025)Zhang, Zhou, Wang, Ji, Huang, and Chen]{zhang2025co}
Xinyu Zhang, Zewei Zhou, Zhaoyi Wang, Yangjie Ji, Yanjun Huang, and Hong Chen.
\newblock Co-mtp: A cooperative trajectory prediction framework with multi-temporal fusion for autonomous driving.
\newblock \emph{arXiv preprint arXiv:2502.16589}, 2025.

\bibitem[Zhao et~al.(2024)Zhao, Xiang, Xu, Xia, Zhou, and Ma]{coopre}
Seth~Z. Zhao, Hao Xiang, Chenfeng Xu, Xin Xia, Bolei Zhou, and Jiaqi Ma.
\newblock Coopre: Cooperative pretraining for v2x cooperative perception, 2024.

\bibitem[Zheng et~al.(2024)Zheng, Xia, Gao, Xiang, and Ma]{cooperfuse}
Zhaoliang Zheng, Xin Xia, Letian Gao, Hao Xiang, and Jiaqi Ma.
\newblock Cooperfuse: A real-time cooperative perception fusion framework.
\newblock In \emph{2024 IEEE Intelligent Vehicles Symposium (IV)}, pages 533--538, 2024.

\bibitem[Zhou et~al.(2022)Zhou, Yang, Zhang, Huang, Chen, and Yu]{zhou_comprehensive_2022}
Zewei Zhou, Ziru Yang, Yuanjian Zhang, Yanjun Huang, Hong Chen, and Zhuoping Yu.
\newblock A comprehensive study of speed prediction in transportation system: {From} vehicle to traffic.
\newblock \emph{iScience}, 25\penalty0 (3):\penalty0 103909, 2022.

\bibitem[Zimmer et~al.(2024)Zimmer, Wardana, Sritharan, Zhou, Song, and Knoll]{zimmer2024tumtraf}
Walter Zimmer, Gerhard~Arya Wardana, Suren Sritharan, Xingcheng Zhou, Rui Song, and Alois~C Knoll.
\newblock Tumtraf v2x cooperative perception dataset.
\newblock In \emph{Proceedings of the IEEE/CVF Conference on Computer Vision and Pattern Recognition}, pages 22668--22677, 2024.

\end{thebibliography}
}

\addtocontents{toc}{\protect\setcounter{tocdepth}{2}} 
\clearpage
\maketitlesupplementary
\renewcommand{\thesection}{\Alph{section}}
\setcounter{section}{0}

\setcounter{table}{0}
\renewcommand{\thetable}{S\arabic{table}}

\setcounter{figure}{0}
\renewcommand{\thefigure}{S\arabic{figure}}

\setcounter{equation}{0}
\renewcommand{\theequation}{S\arabic{equation}}

\localtableofcontents
\vspace{0.5cm}

\section{Implementation Details}
\label{app:implementation}
In this section, we provide detailed configurations for cooperative perception and prediction tasks, including the baseline models used in our experiments and the proposed V2XPnP framework.

\subsection{Benchmark Model Details}
\noindent \textbf{PointPillar Backbone.} 
For all experiments, we employ the anchor-based PointPillar model \cite{lang_pointpillars_2019} as the LiDAR Feature Extraction backbone. The voxel resolution is set to 0.4 meters in both the $x$ and $y$ directions, with a maximum of 32 points per voxel and a total of 32,000 voxels. Additionally, we set the number of anchors per grid cell to 2.

\noindent \textbf{Intermediate Fusion Methods.} 
We implement several state-of-the-art single-frame intermediate fusion methods, including \textit{V2VNet} \cite{wang_v2vnet_2020}, \textit{F-Cooper} \cite{chen2019f}, \textit{DiscoNet} \cite{disconet},\textit{CoBEVFlow} \cite{wei2023asynchronyrobust}, \textit{FFNet} \cite{ffnet}, \textit{V2X-ViT} \cite{xu_v2x-vit_2022}, and our proposed V2XPnP model, integrating them with our end-to-end model to replace the spatio-temporal fusion module. The model settings and configurations for the fusion module adhere to the original implementations.

\noindent \textbf{Map Feature Extraction.} 
HD maps are represented as sets of polylines, with each polyline comprising 10 points. Because the map is projected onto the BEV space, each grid only contains the five nearest polylines. Each waypoint in a polyline contains seven attributes: $(x, y, d_x, d_y, type, x_{pre}, y_{pre})$, representing position, direction, lane type, and previous position. These attributes are encoded using MLP layers into a $256$ hidden dimension feature, followed by 1 or 2 Transformer layers with two attention heads to model interactions among map elements.

\noindent \textbf{Decoupled Attention Predictor.} 
For the decoupled perception and prediction pipeline, we implement an attention-based predictor for trajectory-level prediction tasks. This predictor utilizes a 1D Convolution + LSTM Network \cite{deo_convolutional_2018} to encode temporal historical trajectories and a Transformer layer to capture the interaction among objects and the map, then an LSTM-based decoder generates the future predicted trajectories. All trajectory data, including historical and predicted trajectories, are represented in the local coordinate frame of each object.

\subsection{V2XPnP Model Details}
\noindent \textbf{Temporal Attention.}
To capture the temporal dependence, we initialize the historical timestamp sequence using Sinusoidal positional encodings conditioned on time and further process these encodings through a Linear layer. The temporal attention block in the multi-frame temporal fusion module has four attention heads. To enhance the inter-frame feature representation, we stack three temporal fusion modules with the temporal attention block.

\noindent \textbf{Self-spatial Attention.} 
This block is applied following either the temporal attention or the multi-agent spatial attention. In self-spatial attention, the feature map is partitioned into patches using common window sizes of $(2, 4, 8)$. Given the complexity of spatio-temporal fusion across multiple agents, the self-spatial attention module employs a higher number of attention heads $(16, 8, 4)$ after multi-agent spatial fusion, compared to the heads $(8, 4, 2)$ used following temporal attention. 

\noindent \textbf{Multi-agent Spatial Attention.} 
Our dataset categorizes agents as infrastructure agents, denoted by negative labels (\ie, $-1$ and $-2$), or connected automated vehicles (CAV) agents, denoted by positive labels (\ie, $1$ and $2$). To capture the heterogeneous dependencies among these agents, we construct a heterogeneous graph and employ distinct attention fusion parameters for each agent type. The multi-agent spatial attention utilizes eight attention heads, and we stack three multi-agent spatial fusion modules with the multi-agent spatial attention to capture the inter-agent relationships.

\subsection{Loss Function} 
This section provides the loss function employed in our multi-task model. The initial weights of regression loss $\mathcal{L}_{\text {reg }}$, classification loss $\mathcal{L}_{\text{cla}}$ and prediction loss $\mathcal{L}_{\text{pred}}$ are set as $w_{\text{reg}}, w_{\text{cla}}, w_{\text{pred}} = 2.0, 1.0, 2.0$. For single-task learning, the same loss function is used but weights exclusively on the components relevant to that task.

\noindent \textbf{Perception Loss.} 
The perception task loss combines classification and regression components, designed to align predicted anchor boxes with ground truth labels. For classification, which involves identifying objects and background elements, we employ Focal Loss \cite{lin_focal_2017} to address the imbalance between foreground and background samples. The Focal Loss is expressed as:
\begin{equation}
\mathcal{L}_{\text{cla}} = -\alpha (1 - p_t)^\gamma \log(p_t),
\end{equation}
where $p_t$ is the predicted probability for the target anchor box, and $\alpha$ and $\gamma$ are balancing and focusing factors. Anchor-wise weights are applied to further enhance the balance between positive and negative samples.

For the regression component, we employ Smooth $\ell_1$-Loss to optimize the predicted bounding boxes to match the ground truth labels in terms of position and orientation, and a sine-cosine encoding is employed to handle rotational ambiguities. The Smooth $\ell_1$-Loss is defined as:
\begin{equation}
\mathcal{L}_{\text{reg}} = 
\begin{cases} 
0.5 \cdot \frac{\Delta^2}{\beta}, & \text{if} |\Delta| < \beta, \\
|\Delta| - 0.5 \cdot \beta, & \text{otherwise},
\end{cases}
\end{equation}
where $\Delta=\text{prediction} - \text{target}$, and $\beta$ is a hyper-parameter controlling the transition between $\ell_1$ and $\ell_2$ loss.

\noindent \textbf{Prediction Loss.} We adopt the $\ell_2$-loss function to minimize the discrepancy between the predicted trajectory and the ground truth.
\begin{equation} 
\mathcal{L}_{\text{pred}} = \frac{1}{N_{\text{det}}} \frac{1}{T_{\text{valid}}} \sum_{i=1}^{N_{\text{det}}} \sum_{t=1}^{T_{\text{valid}}} \| \boldsymbol{\mu}^i_t - \mathbf{x}^i_t \|^2, 
\end{equation}
where $\boldsymbol{\mu}^i_t$ and $\mathbf{x}^i_t$ represent the predicted position and target position of the $i$-th object at time step $t$. ${T_{\text{valid}}}$ is the number of valid future time steps for the agent, and $N_{\text{det}}$ is the number of detected objects.

\subsection{Training Strategy}
The end-to-end cooperative perception and prediction model addresses two distinct yet interrelated tasks while integrating information across both temporal and spatial dimensions. Training such an end-to-end model from scratch often results in suboptimal performance, due to the inherent complexity of jointly optimizing these tasks and dimensions. To effectively handle these challenges, we adopt a multi-stage training strategy to progressively refine the model's capabilities. 

\noindent \textbf{Multi-Stage Training Strategy.} 
Initially, the end-to-end perception and prediction model is trained in a single-agent setting, focusing on temporal fusion without incorporating multi-agent spatial fusion. It simplifies the optimization process, enabling the model to learn robust temporal features in isolation. The resulting single-agent model then serves as a pre-trained model for subsequent multi-agent spatial fusion training in the V2X environment. This staged training strategy ensures that the model incrementally acquires the ability to handle the complexities of cooperative perception and prediction tasks.

\noindent  \textbf{Stage 1: Single-Agent Multi-task Learning.} 
The single-agent model training stage addresses the core challenge of coordinating multi-task learning to capture complex patterns across perception and prediction tasks. Prediction task requires a comprehensive understanding of objects' temporal information and their intricate motion patterns, while detection focuses mainly on identifying objects in the current frame, with historical information providing supplementary context. Training both tasks jointly without proper initialization risks overfitting to simpler current-frame features, thereby neglecting the rich but complex temporal features essential for accurate prediction. Moreover, perception is foundational to prediction, as detecting an object is a prerequisite for predicting its motion. To effectively balance the two tasks, we adopt a task-specific training strategy. \textit{(1) Single-Frame Perception Training:} the training begins by optimizing the model for single-frame perception, establishing a foundation for object detection. \textit{(2) Temporal Prediction Training:} the prediction task is introduced by freezing the parameters of the detection backbone and training an additional temporal network and prediction head, guiding the model to focus more on the prediction task and effectively learn complex temporal dependencies. \textit{(3) Joint Fine-Tuning:} the entire model is unfrozen, enabling end-to-end fine-tuning across both tasks.

\begin{table*}[t]
\centering
\caption{Additional benchmark results of cooperative perception and prediction models on V2XPnP Sequential (V2XPnP-Seq) Dataset}
\label{tab:Additional benchmarks}
\vspace{-0.1cm}
\footnotesize
\renewcommand{\arraystretch}{1.15}
\setlength{\tabcolsep}{8.2pt}
\begin{tabularx}{0.95\textwidth}{@{}l|l|cc|c|ccc|c@{}}
\toprule
Dataset & Method & E2E & Map & \textbf{AP@0.5} (\%) $\uparrow$ & ADE (m) $\downarrow$ & FDE (m) $\downarrow$ & MR (\%) $\downarrow$ & \textbf{EPA} (\%) $\uparrow$ \\ \midrule
\multirow{4}{*}{\begin{tabular}[c]{@{}l@{}}V2XPnP-Seq-VC\\ {\scriptsize \textit{(with V+2I at most)}}\end{tabular}}  
    & V2VNet* \cite{wang_v2vnet_2020} & \checkmark & &48.6 &2.10  &3.75  &42.3  &25.3   \\
    & F-Cooper* \cite{chen2019f} & \checkmark & \checkmark &66.0  &1.35  &2.56  &36.1  &38.7  \\
    & DiscoNet* \cite{disconet} & \checkmark & \checkmark &66.8 &1.41  &2.62  &34.4  &42.8   \\
    & \cellcolor{gray!20}V2XPnP (Ours) & \cellcolor{gray!20}\checkmark & \cellcolor{gray!20}\checkmark & \cellcolor{gray!20} \textbf{71.6} & \cellcolor{gray!20} {1.35} &\cellcolor{gray!20} {2.36} &\cellcolor{gray!20} {31.7} &\cellcolor{gray!20} \textbf{48.2} \\ \midrule

\multirow{4}{*}{\begin{tabular}[c]{@{}l@{}}V2XPnP-Seq-IC\\ {\scriptsize \textit{(with 2V+I at most)}}\end{tabular}}  
    & V2VNet* \cite{wang_v2vnet_2020} & \checkmark & &33.6  &1.95  &3.53  &44.2  &16.3  \\
    & F-Cooper* \cite{chen2019f} & \checkmark & \checkmark &60.2  &1.21  &2.32  &36.3  &36.3  \\
    & DiscoNet* \cite{disconet} & \checkmark & \checkmark &65.4 &1.14  &2.18  &36.1  &40.7   \\
    & \cellcolor{gray!20}V2XPnP (Ours) & \cellcolor{gray!20}\checkmark & \cellcolor{gray!20}\checkmark & \cellcolor{gray!20}\textbf{71.0} & \cellcolor{gray!20}1.18 & \cellcolor{gray!20}{2.16} & \cellcolor{gray!20}{34.0} & \cellcolor{gray!20}\textbf{46.0} \\ \midrule

\multirow{4}{*}{V2XPnP-Seq-V2V}  
    & V2VNet* \cite{wang_v2vnet_2020} & \checkmark & &43.1  &3.10  &5.55  &46.8  &19.4  \\
    & F-Cooper* \cite{chen2019f} & \checkmark & \checkmark &60.2  &1.69  &3.22  &41.1  &34.4  \\
    & DiscoNet* \cite{disconet} & \checkmark & \checkmark &61.2 &1.66  &3.13  &41.2  &33.1   \\
    & \cellcolor{gray!20}V2XPnP (Ours) & \cellcolor{gray!20}\checkmark & \cellcolor{gray!20}\checkmark & \cellcolor{gray!20}\textbf{70.5} & \cellcolor{gray!20}{1.78} & \cellcolor{gray!20}{3.28} & \cellcolor{gray!20}{39.9} & \cellcolor{gray!20}\textbf{40.6}  \\ \midrule

\multirow{4}{*}{V2XPnP-Seq-I2I}  
    & V2VNet* \cite{wang_v2vnet_2020} & \checkmark & &41.1  &1.83  &3.34  &40.4  &23.2  \\
    & F-Cooper* \cite{chen2019f} & \checkmark & \checkmark &58.6  &1.34  &2.58  &40.0  &33.6  \\
    & DiscoNet* \cite{disconet} & \checkmark & \checkmark &63.5 &1.15  &2.19  &37.5  &38.4   \\
    & \cellcolor{gray!20}V2XPnP (Ours) & \cellcolor{gray!20}\checkmark & \cellcolor{gray!20}\checkmark & \cellcolor{gray!20}\textbf{69.2}  & \cellcolor{gray!20}{1.26} & \cellcolor{gray!20}{2.31} & \cellcolor{gray!20}{36.5} & \cellcolor{gray!20}\textbf{42.8} \\ \bottomrule
\end{tabularx}%
\vspace{-0.3cm}
\end{table*}

\noindent \textbf{Stage 2: Multi-Agent Spatiotemporal Learning.} 
Based on the pre-trained single-agent model, the multi-agent fusion module is introduced and jointly trained with the entire model. At this stage, the primary focus is to balance the two tasks, ensuring that neither perception nor prediction dominates the training process. To achieve this, we employ a dynamic loss-weighting strategy that gradually increases the weight assigned to the prediction loss. This approach ensures balanced optimization, avoiding performance trade-offs between tasks and improving overall effectiveness across both perception and prediction objectives.

\noindent \textbf{Training Details}. 
The model is trained using the Adam optimizer \cite{kingma2014adam}  with an initial learning rate of $2\times10^{-3}$ and a weight decay of $1\times10^{-4}$ with early stopping on NVIDIA L40S GPUs. We employ $4$ training stages, as detailed before, and each training stage consists of $30$ epochs with a batch size of $2$. Early stopping is employed to prevent overfitting. We carefully tune the hyperparameters to ensure the stability and efficiency of the training process.

\begin{table*}[t]
\caption{Benchmark results for cooperative temporal perception. No Temp: single-frame perception, FaF$^*$: temporal perception with alternating 2D and 3D convolutions, V2XPnP: temporal perception with temporal attention modules.}
\centering
\setlength{\tabcolsep}{6pt} 
\renewcommand{\arraystretch}{1.35} 
\footnotesize
\begin{tabularx}{0.95\textwidth}{Y|ccc|ccc|ccc}
\toprule
\multirow{2}{*}{Dataset} & \multicolumn{3}{c|}{No Fusion (AP@0.5 (\%) $\uparrow$)} & \multicolumn{3}{c|}{Early Fusion (AP@0.5 (\%) $\uparrow$)} & \multicolumn{3}{c}{Intermediate Fusion (AP@0.5 (\%) $\uparrow$)} \\ \cmidrule(l){2-10} 
                         & No Temp   & FaF$^*$   & \cellcolor{gray!20} V2XPnP  & No Temp    & FaF$^*$    & \cellcolor{gray!20} V2XPnP   & No Temp      & FaF$^*$      & \cellcolor{gray!20} V2XPnP     \\ \midrule
V2XPnP-Seq-VC  & 43.9  & 57.1 &\cellcolor{gray!20} \textbf{60.3}  & 63.5  & 67.0  &\cellcolor{gray!20} \textbf{71.0}  & 65.1  & 70.3 &\cellcolor{gray!20} \textbf{74.0} \\ \midrule
V2XPnP-Seq-IC  & 46.4  &61.1  &\cellcolor{gray!20} \textbf{64.7}  & 61.0  & 65.5  &\cellcolor{gray!20} \textbf{71.4}  &61.1  &67.1  &\cellcolor{gray!20} \textbf{73.2}  \\ \midrule
V2XPnP-Seq-V2V & 40.8 &53.7  &\cellcolor{gray!20} \textbf{59.1}  &54.9  &56.4   &\cellcolor{gray!20} \textbf{66.6} &58.0  & 61.4  &\cellcolor{gray!20} \textbf{69.4}   \\ \midrule
V2XPnP-Seq-I2I & 51.0  &61.2  &\cellcolor{gray!20} \textbf{64.7} & 63.4  &66.0  &\cellcolor{gray!20} \textbf{71.6}  &58.5  & 62.9  &\cellcolor{gray!20} \textbf{72.4}  \\ \bottomrule
\end{tabularx}
\label{tab:temporal perception}
\end{table*}

\section{Additional Benchmark Results}
In this paper, we benchmark different spatiotemporal strategies with 11 fusion models in total: 

\begin{itemize}
    \setlength{\itemsep}{3pt}
    \item \textbf{No Fusion}: \textit{No Fusion}, \textit{No Fusion-FaF}
    \item \textbf{Early Fusion}: \textit{Early Fusion}
    \item \textbf{Late Fusion}: \textit{Late Fusion}
    \item \textbf{Intermediate Fusion}: \textit{V2VNet} \cite{wang_v2vnet_2020}, \textit{F-Cooper} \cite{chen2019f}, \textit{DiscoNet} \cite{disconet}, \textit{CoBEVFlow} \cite{wei2023asynchronyrobust}, \textit{FFNet} \cite{ffnet}, \textit{V2X-ViT} \cite{xu_v2x-vit_2022}, and our proposed \textit{V2XPnP}.
\end{itemize}

We present additional benchmark results for \textit{V2VNet} \cite{wang_v2vnet_2020}, \textit{F-Cooper} \cite{chen2019f}, and \textit{DiscoNet} \cite{disconet} across all collaboration modes, as shown in \cref{tab:Additional benchmarks}. Our proposed \textit{V2XPnP} consistently outperforms these SOAT baselines in terms of EPA and AP across all collaboration modes. Notably, V2VNet* exhibits lower performance, likely due to the absence of a map and the loss of temporal features during explicit feature ROI matching.

\section{Cooperative Temporal Perception Task}
In addition to the end-to-end perception and prediction task, the sequential nature of our V2XPnP-Sequential dataset facilitates other temporal tasks, including temporal perception and traditional prediction tasks. In this section, we introduce the cooperative temporal perception task and present benchmark results on the V2XPnP-Sequential dataset. Details on the traditional prediction task are provided in \cref{supp:traditional_prediction}.

\subsection{Problem Formulation}
The cooperative temporal perception task is an extension of the single-frame perception task by incorporating historical context. Specifically, given historical $T$ frames raw perception data ${\mathbf{P}_i^t}, i\in\{1,\cdots,N\}$ from all $N$ agents within the communication range of the ego agent, the objective is to detect the surrounding objects in the current frame. The core challenge lies in effectively leveraging temporal information from $T$ past frames to enhance detection accuracy in the present frame.

\subsection{Benchmark Methods}
For benchmarking, we adapt our end-to-end model, \textit{V2XPnP}, by removing the prediction head, resulting in a model only for temporal perception. Various V2X fusion strategies are evaluated in this framework, as detailed in \cref{tab:temporal perception}.  Moreover, we provide another baseline \textit{FaF$^*$}, which adopts a combination of 2D and 3D convolutions for temporal fusion. \textit{FaF$^*$} further integrates with the F-Cooper intermediate fusion method and early fusion method for V2X fusion comparison. We also provide the results of the \textit{No Temp} model, which excludes temporal fusion and is evaluated using both F-Cooper and early fusion methods. Model parameters and experimental setups for this task are consistent with those used for the end-to-end cooperative perception and prediction task.

\subsection{Benchmark Results}
The results demonstrate that incorporating temporal cues significantly improves perception performance across all multi-agent fusion strategies. Notably, our \textit{V2XPnP} model achieves superior results compared to other baselines, due to the careful design of temporal attention. However, we observe a slight performance drop when the same model is applied to the end-to-end cooperative perception and prediction task, compared to its use solely for temporal perception. The possible reason is the difficulty of optimizing both tasks to achieve optimal performance. Nevertheless, the end-to-end model still outperforms other baselines in both perception and prediction tasks. Future research should focus on optimizing the balance between multiple tasks to further enhance the performance of end-to-end models.

\begin{table*}[t]
\caption{Benchmark results for traditional prediction. No Fusion: prediction based on the no-fusion perception results. Late Fusion: prediction based on the late fusion perception results. Ground Truth: prediction based on the ground truth trajectories with no occlusion or perception errors.}
\centering
\setlength{\tabcolsep}{5pt} 
\renewcommand{\arraystretch}{1.15} 
\footnotesize
\begin{tabularx}{0.95\textwidth}{@{}Y|l|cccc|cccc@{}}
\toprule
\multirow{2}{*}{Dataset} &
  \multirow{2}{*}{Method} &
  \multicolumn{4}{c|}{Attention Predictor} &
  \multicolumn{4}{c}{LSTM Predictor} \\ \cmidrule(l){3-10} 
 &
   &
  \multicolumn{1}{c|}{AP@0.5(\%) $\uparrow$} &
  ADE(m) $\downarrow$ &
  FDE(m) $\downarrow$ &
  MR(\%) $\downarrow$ &
  \multicolumn{1}{l|}{AP@0.5(\%) $\uparrow$} &
  ADE(m) $\downarrow$ &
  FDE(m) $\downarrow$ &
  MR(\%) $\downarrow$ \\ \midrule
\multicolumn{1}{l|}{\multirow{3}{*}{\begin{tabular}[c]{@{}l@{}}V2XPnP-Seq\\ -VC\end{tabular}}} &
  No Fusion &
  \multicolumn{1}{c|}{43.9} &
  1.87 &
  3.24 &
  33.8 &
  \multicolumn{1}{c|}{43.9} & 
   2.91 &
   4.77 &
   35.0 
   \\
\multicolumn{1}{l|}{} &
  Late Fusion &
  \multicolumn{1}{c|}{58.1} &
  1.59 &
  2.81 &
  34.3 &
  \multicolumn{1}{c|}{58.1} & 
  2.76 &
  4.60 &
  33.7
   \\
\multicolumn{1}{l|}{} &
  Ground Truth &
  \multicolumn{1}{c|}{-} & 0.60
   & 1.26
   & 23.0
   &
  \multicolumn{1}{c|}{-} & 0.66
   & 1.31
   & 23.0
   \\ \midrule
\multicolumn{1}{l|}{\multirow{3}{*}{\begin{tabular}[c]{@{}l@{}}V2XPnP-Seq\\ -IC\end{tabular}}} &
  No Fusion &
  \multicolumn{1}{c|}{46.4} &
  2.10 &
  3.75 &
  42.3 &
  \multicolumn{1}{c|}{46.4} &
   2.11&
   3.67&
   35.8\\
\multicolumn{1}{l|}{} &
  Late Fusion &
  \multicolumn{1}{c|}{55.9} &
  1.39 &
  2.44 &
  30.1 &
  \multicolumn{1}{c|}{55.9} &
   2.61&
   4.40&
   32.7\\
\multicolumn{1}{l|}{} &
  Ground Truth &
  \multicolumn{1}{c|}{-} &
   0.63&
   1.35&
   26.2&
  \multicolumn{1}{c|}{-} &
   0.61&
   1.31&
   25.0 \\ \midrule
\multicolumn{1}{l|}{\multirow{3}{*}{\begin{tabular}[c]{@{}l@{}}V2XPnP-Seq\\ -V2V\end{tabular}}} &
  No Fusion &
  \multicolumn{1}{c|}{40.8} &
  1.99 &
  3.38 &
  34.0 &
  \multicolumn{1}{c|}{40.8} &
   2.98&
   4.82&
   34.4\\
\multicolumn{1}{l|}{} &
  Late Fusion &
  \multicolumn{1}{c|}{55.3} &
  1.75 &
  3.07 &
  34.0 &
  \multicolumn{1}{c|}{55.3} &
   2.87&
   4.79&
   35.0\\
\multicolumn{1}{l|}{} &
  Ground Truth &
  \multicolumn{1}{c|}{-} &
   0.60&
   1.26&
   22.9&
  \multicolumn{1}{c|}{-} &
   0.66&
   1.31&
   22.8\\ \midrule
\multicolumn{1}{l|}{\multirow{3}{*}{\begin{tabular}[c]{@{}l@{}}V2XPnP-Seq\\ -I2I\end{tabular}}} &
  No Fusion &
  \multicolumn{1}{c|}{51.0} &
  1.69 &
  3.06 &
  36.2 &
  \multicolumn{1}{c|}{51.0} &
   2.11&
   3.67&
   35.9\\
\multicolumn{1}{l|}{} &
  Late Fusion &
  \multicolumn{1}{c|}{61.3} &
  1.41 &
  2.50 &
  30.0 &
  \multicolumn{1}{c|}{61.3} &
   2.44&
   4.18&
   32.1\\
\multicolumn{1}{l|}{} &
  Ground Truth &
  \multicolumn{1}{c|}{-} &
   0.63&
   1.35&
   26.2&
  \multicolumn{1}{c|}{-} &
   0.61&
   1.31&
   25.0\\ \bottomrule
\end{tabularx}
\label{tab:prediction}
\end{table*}

\section{Traditional Cooperative Prediction Task}
\label{supp:traditional_prediction}

\subsection{Problem Formulation}
V2XPnP sequential dataset also supports the traditional prediction task. Compared to end-to-end models, which directly infer future states of objects from perception data, the traditional prediction task forecasts their future trajectories from historical trajectories. The cooperative prediction task is formulated as follows: given the map and the historical trajectories of all detected objects obtained from the ego agent and other agents (\eg, CAVs and infrastructure units) within the communication range of the ego agent, the objective is to predict future trajectories of these detected objects.

\subsection{Benchmark Methods}
To investigate the influence of perception results on prediction tasks, we provide two types of input for the prediction models: 1) Ground-truth historical trajectories of surrounding objects; 2) Perception-based historical trajectories generated by the upstream perception module. The first one is the common setting for the traditional trajectory prediction task, assuming full availability of accurate historical trajectories for prediction. However, it ignores real-world challenges such as occlusions and cumulative errors introduced by separate modules. To address this limitation and enable a more realistic evaluation, we designed the second setting, where CAVs can only derive the historical trajectories from the perception results, and thus the perception uncertainty can propagate to the downstream prediction. Notably, regardless of the input type, the prediction model is trained using the complete future trajectory dataset aggregated from all agents.

In our experiment, the prediction model configuration and experimental setup align closely with the decoupled attention predictor. Following the LSTM baseline setting in the Waymo motion dataset \cite{ettinger2021large}, the LSTM model also serves as a strong baseline, which includes an LSTM encoder and LSTM decoder.  We report benchmark results under three configurations: \textit{No Fusion}, where no perception information is fused; \textit{Ground Truth}, assuming perfect historical trajectories; \textit{Late Fusion}, where the decoupled pipeline from the traditional prediction task is employed.

\subsection{Benchmark Results}
The experimental results, summarized in \cref{tab:prediction}, compare traditional prediction under three input settings: ground truth trajectories, perception without fusion, and perception with late fusion. The results indicate that as perception improves—from no fusion to late fusion—the prediction performance correspondingly increases. When the environment is fully observable, the task simplifies to the traditional prediction setup, achieving the best overall performance for both detection and prediction. A significant drop in performance is observed for perception-based prediction, highlighting the critical dependency of predictive tasks on perception accuracy. Moreover, the Attention predictor shows better robustness compared to the LSTM baseline under noisy perception inputs, thanks to the attention module for complex interaction feature capturing. We anticipate that this temporal prediction task will inspire further exploration of perception-based prediction approaches.

\begin{figure*}[t]
    \vspace{2cm}
    \centering
    \includegraphics[width=\textwidth]{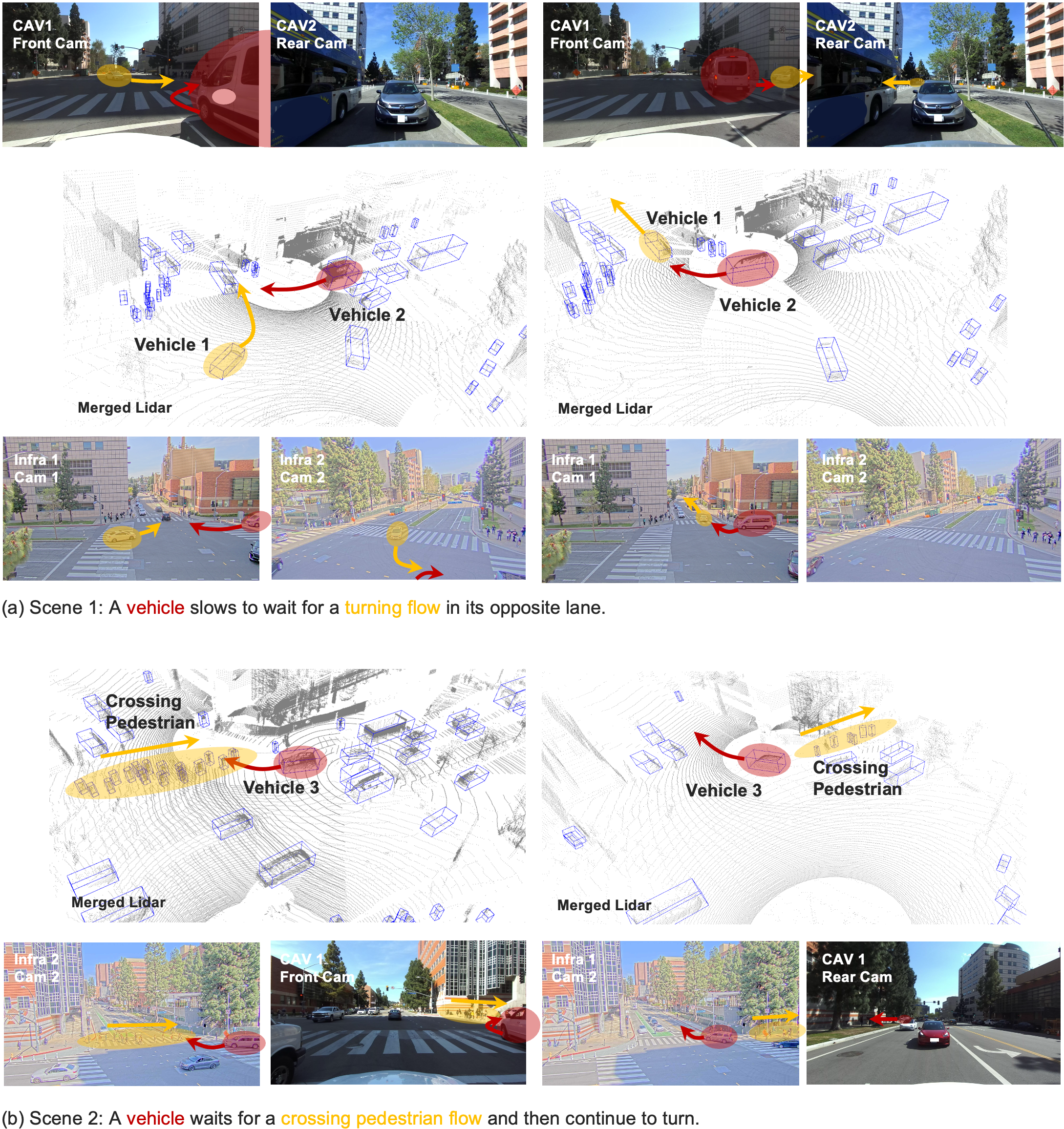}
    \caption{Examples of interaction scenarios from the V2XPnP-Sequential dataset. The dataset provides multi-agent perception perspectives and captures diverse interaction behaviors among ten object classes in dense urban traffic environments.}
    \label{fig:interaction_samples}
\end{figure*}

\section{V2XPnP Sequential Dataset Details}
\label{supp:dataset_details}

\subsection{Dataset Visualization}
Our V2XPnP-Sequential dataset provides two sensor sequences (LiDAR and camera) collected in dense urban environments, capturing diverse interactive behaviors over time. \cref{fig:interaction_samples} illustrates two representative interaction scenarios in our dataset, presenting LiDAR and camera data from multiple agents at two key timestamps. The main intersection objects pair has been annotated with \textcolor{myred}{red} and \textcolor{myyellow}{yellow} blocks in different agents' views. 

\subsection{Data Acquisition}
\noindent \textbf{Sensor Specifications.} 
The dataset was collected using four agents - two connected automated vehicles and two smart infrastructure units. Each CAV is equipped with a RoboSense 128-beam LiDAR, four stereo RGB cameras with 1920 × 1080 resolution, and an integrated GPS/IMU system. The four stereo cameras are mounted on the front, rear, left, and right sides of the CAV, providing a complete 360-degree field of view. Similarly, each infrastructure unit is configured with a 128- or 64-beam LiDAR, two Axis cameras with 1920 × 1080 resolution, and a GPS module. The sensor deployment of our data collection system is shown in Fig. 4(a).

\begin{table*}[t]
\centering
\caption{Comparison between the V2XPnP-Sequential dataset and other public available driving datasets}
\renewcommand{\arraystretch}{1.15} 
\label{tab:dataset}
\setlength{\tabcolsep}{5.2pt} 
\footnotesize
\begin{tabularx}{\textwidth}{l|c|c|ccc|cc|c|cc|cc|c}
\toprule
Dataset &
  Year &
  Type &
  V2V &
  V2I &
  I2I &
  Trajectory &
  Map &
  \begin{tabular}[c]{@{}c@{}}Agent \\ Number\end{tabular} &
  \begin{tabular}[c]{@{}c@{}}Tracked \\ Objects/Scene\end{tabular} &
  \begin{tabular}[c]{@{}c@{}}3D \\ Boxes\end{tabular} &
  \begin{tabular}[c]{@{}c@{}}RGB \\ Images\end{tabular} &
  \begin{tabular}[c]{@{}c@{}}LiDAR \\ Frames\end{tabular} &
  Categories \\ \midrule
nuScenes \cite{caesar_nuscenes_2020}      & 2019 & Real &            &            &            & \checkmark & \checkmark & 1    & 75.75 & 1.4M  & 1.4M & 400k & 23 \\
Waymo Open \cite{ettinger2021large}    & 2019 & Real &            &            &            & \checkmark & \checkmark & 1    & -     & 12M   & 1M   & 200k & 4  \\ \midrule
OPV2V \cite{xu_opv2v_2022}         & 2022 & Sim  & \checkmark &            &            &            &            & 2.89 & 26.5  & 230k  & 44k  & 11k  & 1  \\
V2X-Sim \cite{li_v2x-sim_2022}       & 2022 & Sim  & \checkmark & \checkmark &            & \checkmark &            & 10   & -     & 26.6k & 0    & 10k  & 1  \\
V2XSet \cite{xu_v2x-vit_2022}        & 2022 & Sim  & \checkmark & \checkmark &            & \checkmark &            & 2-7  & -     & 230k  & 44k  & 11k  & 1  \\ \midrule
DAIR-V2X \cite{yu_dair-v2x_2022}      & 2022 & Real &            & \checkmark &            &            &            & 2    & 0     & 464k  & 39k  & 39k  & 10 \\
V2V4Real \cite{xu_v2v4real_2023}      & 2023 & Real & \checkmark &            &            & \checkmark & \checkmark & 2    & -     & 240k  & 40k  & 20k  & 5  \\
V2X-Seq \cite{yu_v2x-seq_2023}       & 2023 & Real &            & \checkmark &            & \checkmark & \checkmark & 2    & 110   & 464k  & 71k  & -    & 10 \\
RCooper \cite{hao2024rcooper}       & 2024 & Real &            &            & \checkmark & \checkmark &            & 4    & -     & -     & 50k  & 30k  & 10 \\
V2X-Real \cite{xiang2024v2x}      & 2024 & Real & \checkmark & \checkmark & \checkmark &            &            & 4    & 0     & 1.2M  & 171K & 33k  & 10 \\ \midrule
\rowcolor{gray!20} \textbf{V2XPnP-Seq} & \textbf{2024} & \textbf{Real} & \textbf{\checkmark} & \textbf{\checkmark} & \textbf{\checkmark} & \textbf{\checkmark} & \textbf{\checkmark} & \textbf{4}    & \textbf{136}   & \textbf{1.45M}  & \textbf{208k} & \textbf{40K} & \textbf{10}  \\ \bottomrule
\end{tabularx}%
\end{table*}

\noindent \textbf{Coordinate System.} 
Our V2XPnP-Sequential dataset encompasses three coordinate systems: the LiDAR coordinate system, the camera coordinate system, and the map coordinate system. Each agent (vehicle or infrastructure) maintains its own local LiDAR and camera coordinate systems. The global map coordinate system serves as the reference for all annotations and maps. The transformation from each agent's local LiDAR coordinate to the map coordinate in each frame is achieved with the GPS/IMU data and the offline PCD map. We also conduct the 3D-2D calibration for LiDAR and camera, as shown in Fig. 4(b). 

\subsection{Data Annotation and Processing}
\noindent \textbf{Data Annotation.} 
The 3D bounding boxes in our V2XPnP-Sequential dataset are annotated using an open-source labeling tool, SUSTechPOINTS \cite{li2020sustech}, by expert annotators. The first step is annotating the bounding boxes in the point clouds from the two CAVs and infrastructure units. Then, these bounding boxes, annotated in different agents’ coordinate frames, are processed through a V2X sequential pipeline to assign consistent object IDs across agents and temporal frames.  To ensure annotation quality, each object is subjected to eight rounds of review and revision. In total, ten object categories are included in our dataset: car, pedestrian, scooter, motorcycle, bicycle, truck, van, concrete truck, bus, and road barrier. Each object annotation includes the center of the bounding box ($x, y, z$), sizes ($width, length, height$), and orientation ($roll, yaw, pitch$) in the global coordinates. Notably, we follow a general object definition in annotation, encompassing stationary objects such as parked vehicles and barriers, which are annotated similarly to movable objects but explicitly labeled as static. This aligns with public datasets like nuScenes \cite{caesar_nuscenes_2020}, where static objects are tracked while maintaining consistent IDs.

\noindent \textbf{Trajectory Generation.} 
In addition to perception data, the dataset provides a ground-truth trajectory dataset derived from the fused perception data of all agents, capturing the trajectories of objects across all frames. This trajectory dataset is primarily utilized in traditional prediction tasks, which assume all history trajectories are observable to the ego agent. However, this assumption ignores the fact that the trajectories obtained from onboard sensors are incomplete due to occlusion and limited perception range, and no specific datasets are designed to support this task. To support research in prediction with real-world sensor constraints, we provide a trajectory retrieval module in the V2XPnP-Sequential dataset to return observable trajectories of surrounding objects based on their actual visibility relationships.

\noindent \textbf{Map Generation.} 
The HD map generation involves two stages: point cloud (PCD) map generation and vector map generation. (1) To generate the PCD map, each LiDAR frame from the CAVs is preprocessed to remove dynamic objects, retaining only static elements essential for mapping. Then, a Normal Distribution Transform (NDT) scan-matching algorithm is employed to compute the relative transformation between consecutive frames, forming the basis of the LiDAR odometry. We also incorporate translation and heading information obtained from the vehicle's GPS/IMU system, integrating them through a Kalman filter to refine the pose estimation, mitigating the drift from the error accumulation in LiDAR data. Finally, the LiDAR sequences are fused to form the PCD map across all collection areas. (2) The aggregated PCD map is imported into RoadRunner \cite{crescenzi2001roadrunner} to generate vector maps. Road geometry is inferred and annotated based on intensity variations visualized by distinct color mappings within RoadRunner, and all the semantic attribution is annotated based on the collected camera data, such as road type (\eg, driving, sidewalk, and parking) and line type (\eg, solid and broken yellow line combination and solid white line). Finally, the generated maps are exported in the OpenDRIVE (Xodr) format and converted to Waymo map format \cite{ettinger2021large}, ensuring compatibility with downstream applications.

\begin{figure}[t]
    \centering
    \includegraphics[width=\columnwidth]{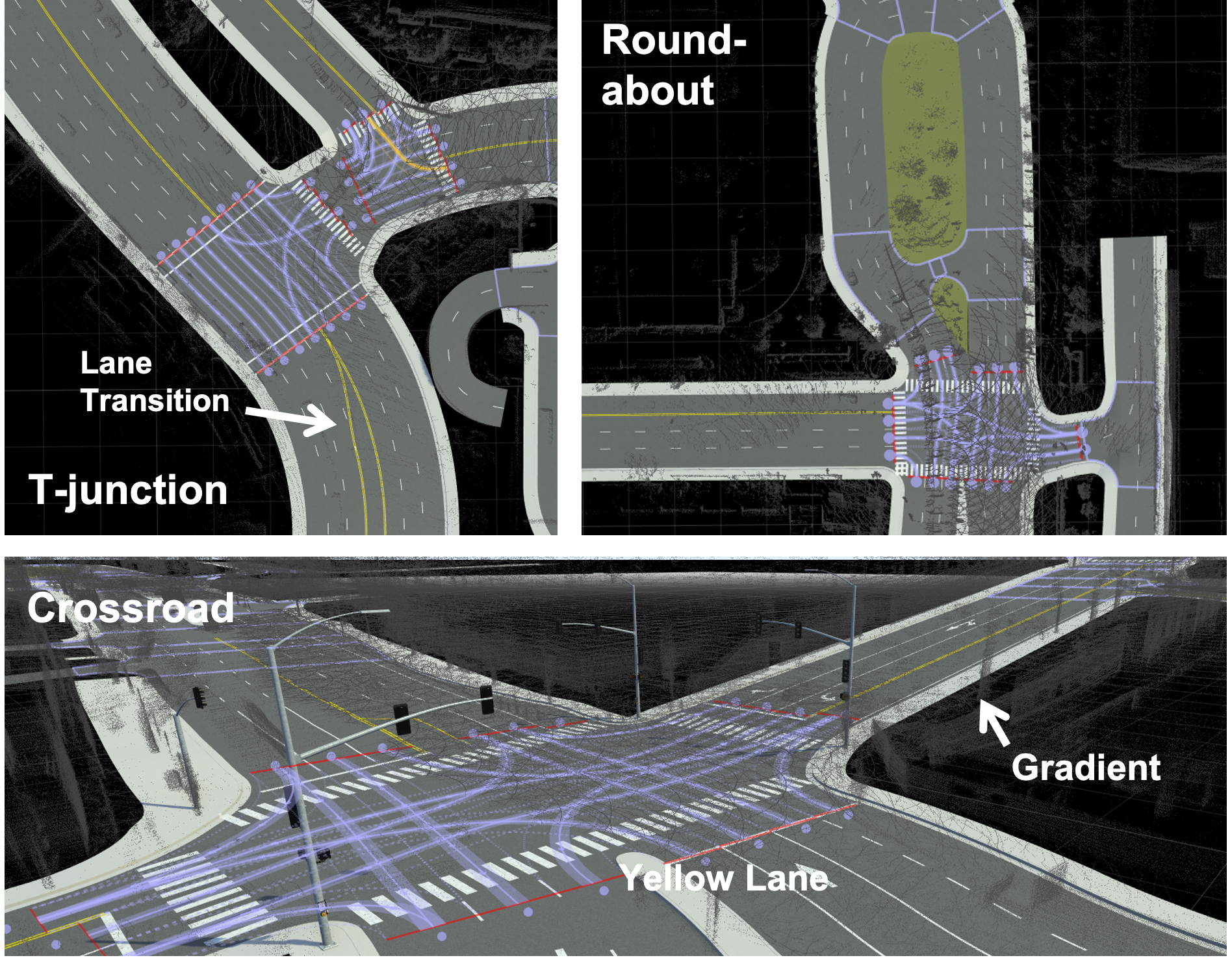}
    \caption{Examples of intersection types in the map, including T-junctions, roundabouts, and crossroads. The gray point clouds in the background represent the PCD map, while lane transitions and gradients are depicted in the map.}
    \label{fig:map}
\end{figure}

\subsection{Dataset Analysis}
\cref{tab:dataset} presents the comparison of the V2XPnP-Sequential dataset with existing driving datasets. Our dataset tracks an average of 136 objects per scene, recording high-density and complex traffic scenarios. Furthermore, the dataset's extensive map and trajectory data further enhance its utility in cooperative perception and prediction research across all collaboration modes. The data distribution of ten object classes is shown in Fig. 4(d). The dataset covers 24 intersections of varying types, including roundabouts, T-junctions, and crossroads, as shown in \cref{fig:map}. Notably, many collection areas have a significant gradient, which can facilitate the detection and prediction research in diverse terrain conditions.

\subsection{Dataset Privacy Protection}
The V2XPnP-Sequential dataset is designed with stringent privacy safeguards to ensure the anonymity of individuals and vehicles. Trajectory data only includes object IDs and positions, eliminating the possibility of tracking specific entities. All perception data undergoes privacy-preserving processing, with LiDAR annotations retaining only essential attributes such as object ID, agent type, and bounding box pose. Additionally, all image data has been anonymized, with human faces and other potentially sensitive details obscured or removed.

\end{document}